\documentclass[pdflatex,sn-mathphys]{sn-jnl}
\jyear{2021}%
\theoremstyle{thmstyleone}%
\theoremstyle{thmstyletwo}%
\theoremstyle{thmstylethree}%
\newcommand\inv[1]{#1\raisebox{1.15ex}{$\scriptscriptstyle-\!1$}}
\usepackage{arydshln}
\usepackage{subfigure}
\newcommand*{\review}{\textcolor{black}}
\newcommand*{\Review}{\textcolor{black}}
\usepackage{comment}
\usepackage{etoolbox}
    \makeatletter
    \patchcmd{\ps@headings}
    {\hbox to \hsize{\hfill Springer \template\hfill}}
    {\hbox to \hsize{\hfill \hfill}}
    {}
    {}
    \patchcmd{\ps@titlepage}
    {\hbox to \hsize{\hfill Springer \template\hfill}}
    {\hbox to \hsize{\hfill \hfill}}
    {}
    {}
    \makeatother
\begin{document}

\title[Deep Transfer Operator Learning]{Deep transfer operator learning for partial differential equations under conditional shift}

\author[1]{\fnm{Somdatta} \sur{Goswami}}\email{somdatta\_goswami@brown.edu}
\equalcont{These authors contributed equally to this work.}

\author[2]{\fnm{Katiana} \sur{Kontolati}}\email{kontolati@jhu.edu}
\equalcont{These authors contributed equally to this work.}

\author*[2]{\fnm{Michael} \sur{D. Shields}}\email{michael.shields@jhu.edu}

\author*[1]{\fnm{George} \sur{Em Karniadakis}}\email{george\_karniadakis@brown.edu}

\affil*[1]{\orgdiv{Division of Applied Mathematics}, \orgname{Brown University}, \orgaddress{\street{170 Hope Street}, \city{Providence}, \postcode{02906}, \state{Rhode Island}, \country{U.S.A}}}

\affil[2]{\orgdiv{Department of Civil and Systems Engineering}, \orgname{Johns Hopkins University}, \orgaddress{\street{3400 N. Charles Street}, \city{Baltimore}, \postcode{21218}, \state{Maryland}, \country{U.S.A}}}

\abstract{
Transfer learning (TL) enables the transfer of knowledge gained in learning to perform one task (source) to a related but different task (target), hence addressing the expense of data acquisition and labeling, potential computational power limitations, and dataset distribution mismatches. We propose a new TL framework for task-specific learning (functional regression in partial differential equations (PDEs)) under conditional shift based on the deep operator network (DeepONet). Task-specific operator learning is accomplished by fine-tuning task-specific layers of the target DeepONet using a hybrid loss function that allows for the matching of individual target samples while also preserving the global properties of the conditional distribution of target data. Inspired by the conditional embedding operator theory, we minimize the statistical distance between labeled target data and the surrogate prediction on unlabeled target data by embedding conditional distributions onto a reproducing kernel Hilbert space. We demonstrate the advantages of our approach for various TL scenarios involving nonlinear PDEs under diverse conditions due to shift in the geometric domain and model dynamics. Our TL framework enables fast and efficient learning of heterogeneous tasks despite significant differences between the source and target domains.}


\maketitle


\section*{Main Text}\label{sec2}

Deep learning has been successfully employed to simulate computationally expensive complex physical processes described by partial differential equations (PDEs) and achieve superior performance that allows the acceleration of numerous tasks including uncertainty quantification (UQ), risk modeling and design optimization \citep{chen2018neural,raissi2019physics,li2021fourier,lu2021learning, chatterjee2021robust, olivier2021bayesian}. 
Despite this success, the predictive performance of such models is often limited by the availability of labeled data used for training. However, in many cases collecting large and sufficient labeled datasets can be computationally intractable (e.g., when high-fidelity or multi-scale models are considered). Furthermore, learning in isolation, i.e., training a single predictive model for different but related tasks, can be extremely expensive. To tackle this bottleneck, knowledge between relevant domains can be leveraged in a framework known as \textit{transfer learning} (TL) \cite{niu2020decade}. In this scenario, information from a model trained on a specific domain (\textit{source}) with sufficient labeled data can be transferred to a different but closely related domain (\textit{target}) for which only a small number of training data is available.

In machine learning, TL is a popular and promising area and has been applied to address the issue of data scarcity in various problems, including image recognition \cite{gao2018deep,yang2021image} and natural language processing (NLP) \cite{ruder2019transfer,zhang2021combining}. The most important prerequisite in TL is that there needs to be a connection between the learning domains. In fact, one proposed categorization of TL approaches is based on the consistency between the distributions of source and the target input (or feature) spaces and output (or label) spaces \cite{zhuang2020comprehensive}. The shift between the source and target data distributions is considered the major challenge in modern TL. The types of distribution shifts include conditional shift, where the marginal distribution of source and target input data remains the same while the conditional distributions of the output differ (i.e., $P(\mathbf{x}_s) = P(\mathbf{x}_t)$ and $P(\mathbf{y}_s\vert\mathbf{x}_s) \ne P(\mathbf{y}_t\vert\mathbf{x}_t)$) and covariate shift, where the opposite occurs (i.e., $P(\mathbf{x}_s) \ne P(\mathbf{x}_t)$ and $P(\mathbf{y}_s\vert\mathbf{x}_s) = P(\mathbf{y}_t\vert\mathbf{x}_t)$). TL for problems under covariate shift have been explored primarily for classification and more recently for regression tasks \cite{certo2016sample,chen2021representation}. State-of-the-art approaches for regression problems under conditional shift include TL by Boosting (TLB) \cite{pardoe2010boosting}, the Residual Approximation (RA) \cite{wang2014active}, General Transformation Function (GTF) \cite{du2017hypothesis} and Domain Adaptation Under GeTarS \cite{zhang2013domain}. Other more recent methods include ResTL, an approach based on fuzzy residuals \cite{chen2020transfer}, and RSD+BMP based on a subspace representation distance and penalization of bases mismatch \cite{liu2021deep}. Zhang and Garikipati proposed \cite{zhang2020machine} Knowledge-Based Neural Networks (KBNNs) that take advantage of the dominant characteristics of the data to study evolving microstructures. Recently, approaches based on Physics-Informed Neural Networks (PINNs) \cite{goswami2020transfer,desai2021one,chen2021transfer,penwarden2021physics, wang2022mosaic} have been proposed to solve PDE problems under distribution shift.

\begin{figure}[ht!]
\begin{center}
\includegraphics[width=1\textwidth]{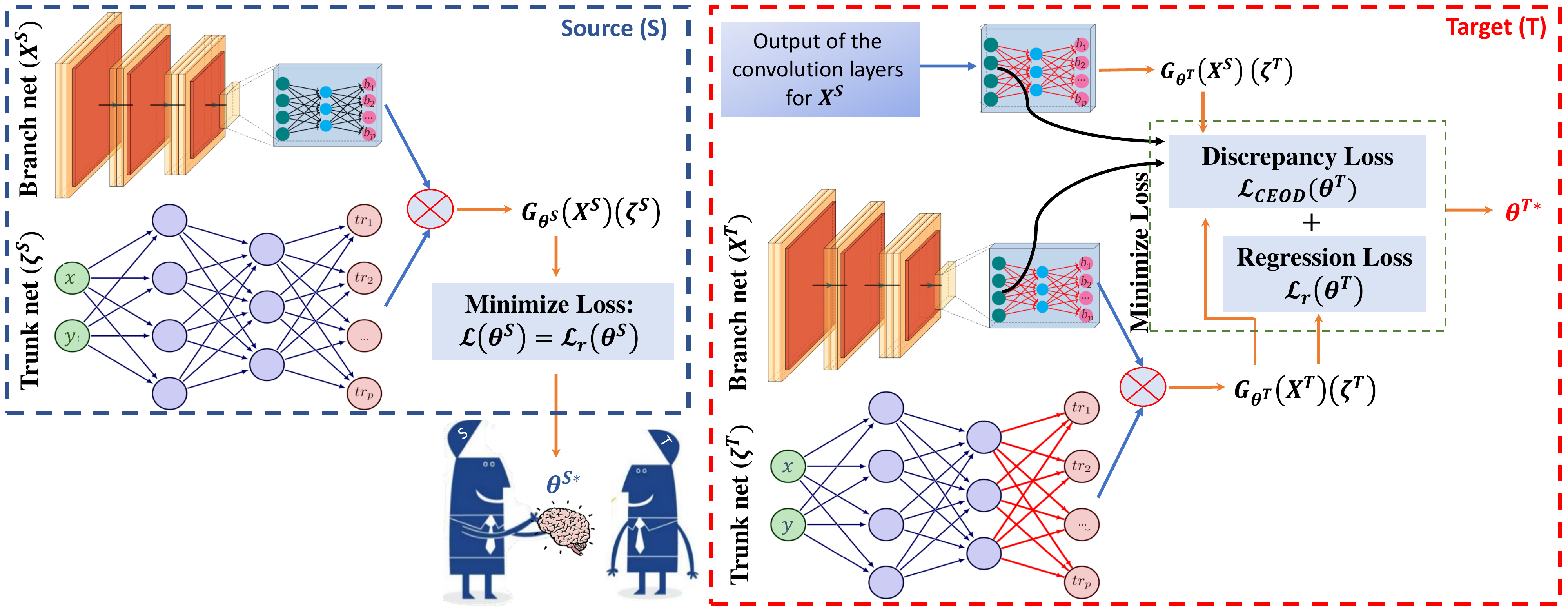}
\caption{The proposed transfer learning framework to approximate PDE solutions using the standard DeepONet. The source model (DeepONet) (blue box), aims to learn the mapping from $\mathbf X$ (input to the branch net) to the solution, employing a solution operator, $\mathcal G_{\theta^S}$, evaluated at $\zeta^S$ locations, where $\theta^S$ denotes the trainable parameters of the source model. The loss function, $\mathcal L(\theta^S)$, obtained as the residual loss, $\mathcal L_r(\theta^S)$, is minimized to obtain the optimized parameters of the network, $\theta^{S*}$. The optimized trainable parameters of the source model are transferred to the target model (TL-DeepONet). All the layers of the target model (red box) are frozen except for the last layer of the trunk net and the fully-connected layers of the convolutional neural network in the branch net (shown with red arrows). The parameters associated with these layers are fine-tuned based on a hybrid loss function $\mathcal L(\theta^T)$, which is minimized to obtain the optimized parameters of the target model, $\theta^{T*}$.}
\label{fig:DeepONet}
\end{center}
\end{figure}

Motivated by the lack of TL approaches for task-specific operator learning and UQ, in this work we present a novel framework for efficient TL under conditional shift using neural operators, see Figure \ref{fig:DeepONet}. We refer to the target neural operator as \review{{\em Transfer Learning Deep Operator Network} or {\em TL-DeepONet}}. The main idea behind this work is to train a source model with sufficient labeled data (i.e., model evaluations) from a source domain under a standard regression loss and transfer the learned variables to a second target model, which is trained with very limited labeled data from a target domain under a hybrid loss function. The hybrid loss is comprised of a regression loss and the conditional embedding operator discrepancy (CEOD) loss \cite{liu2021deep}, used to measure the divergence between conditional distributions in a reproducing kernel Hilbert space (RKHS). The key ingredient of the proposed framework is the exploitation of domain-invariant features extracted by the source model, which leads to the efficient initialization of the target model variables. It is widely accepted that lower layers are in charge of more general features \cite{neyshabur2020being}, thus fine-tuning of the target model focuses on training higher levels of the network. \review{Importantly, fine-tuning of the network is faster than training the entire network from scratch with random initialization, which results in significant computational savings especially when multiple target tasks are considered.}

While most surrogate modeling techniques, such as standard neural networks, aim to simply approximate a mapping representing the solution of a PDE on fixed discretized domains, operator regression methods approximate mappings between infinite-dimensional spaces and are thus discretization-invariant. \review{Some of the operator regression methods are the Fourier neural operator (FNO) \cite{li2021fourier}, wavelet neural operator (WNO) \cite{tripura2022wavelet} and the graph kernel network (GKN) \cite{li2020neural}, to name a few. In particular, FNO has shown promising results in domains with structured meshes in low dimensions, and some significant enhancements to FNO have been proposed to deal with mappings of different dimensionality as well as with complex-geometry domains \cite{lu2021comprehensive}. In this work, we employ a more general deep neural operator (DeepONet) \cite{lu2021learning}, which allows us to fully learn the operator and thus perform real-time prediction for arbitrary new inputs and complex domains.} Importantly, the proposed transfer learning framework enables the identification of the PDE operator in domains where very limited labeled data are available. To the best of our knowledge, this is the first comprehensive study on the implementation of TL for operator learning in PDE problems. A schematic of the proposed approach is presented in Figure \ref{fig:DeepONet}. The main contributions of this work can be summarized as follows: 

\begin{itemize}
    \item We propose a novel framework for transfer learning problems under conditional shift with deep neural operators.
    \item  The proposed framework can be employed for fast and efficient task-specific PDE learning and uncertainty quantification.
    \item We leverage principles of the RKHS and the conditional embedding operator theory to construct a new hybrid loss function and fine-tune the target model.
    \item The \review{advantages and limitations} of the proposed framework are demonstrated through a variety of transfer learning problems, including distribution shifts due to changes in the domain geometry, model dynamics, material properties, non-linearties and more.
\end{itemize}

\begin{figure}[ht!]
\begin{center}
\includegraphics[width=1\textwidth]{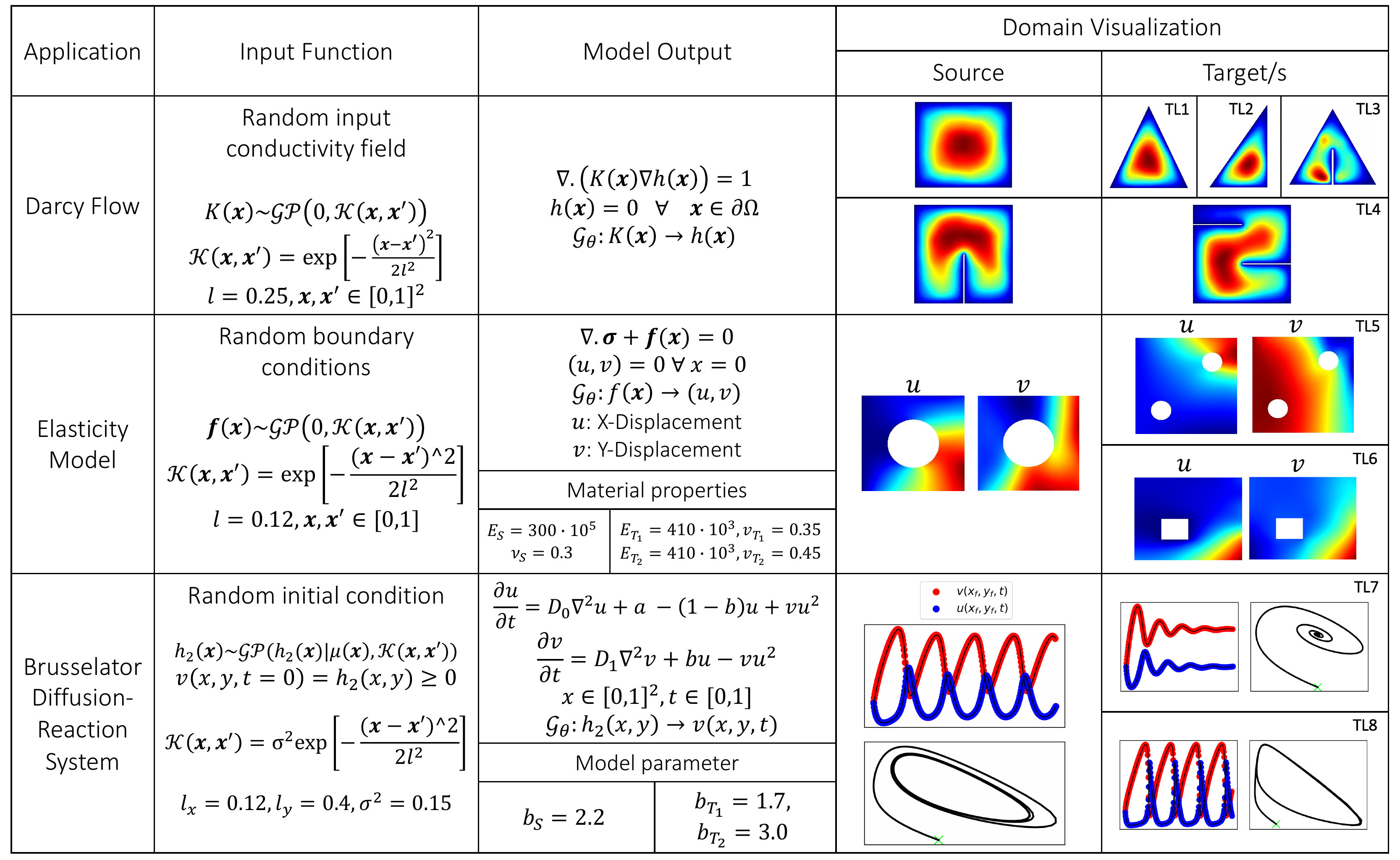}
\caption{\review{A schematic representation of the operator learning benchmarks and TL scenarios under consideration in this work. The input/output functions and representative plots of the source (DeepONet) and the target domains (TL-DeepONet) are shown. In addition, we have simulated five more cases, TL9 - TL13, see SI.}}
\label{fig:applications}
\end{center}
\end{figure}

We present a comprehensive collection of transfer learning problems for parametric PDEs to evaluate the validity of the proposed approach. A visual description of the different benchmarks considered is presented in Figure \ref{fig:applications}. We first introduce the benchmark problems along with the transfer learning scenarios considered and then provide the experimental results. The DeepONet architectures considered in each transfer learning case are shown in Table \ref{table:architectures}.

\bigbreak
\noindent
\textbf{Darcy flow} 

\noindent
Darcy's law describes the pressure of a fluid flowing through a porous medium with a given permeability and can be mathematically expressed by the following system of equations:
\begin{equation}
\label{eq:Darcy}
    \nabla \cdot (K(\boldsymbol x)\nabla h(\boldsymbol x)) = g(\boldsymbol x), \;\;\boldsymbol{x} = (x,y),
\end{equation}
subject to the following boundary conditions
\begin{equation*}
\label{eq:Darcy-bcs}
    h(\boldsymbol x) = 0, \hspace{10pt} \forall \; \boldsymbol x \in \partial \Omega,
\end{equation*}
where $K(\boldsymbol x)$ is the spatially varying hydraulic conductivity of the heterogeneous porous media, and $h(\boldsymbol x)$ is the corresponding hydraulic head. For simplicity, we consider a fixed forcing term, i.e., $g(\boldsymbol x) = 1$. \review{Problems for which practitioners can find transfer learning useful include applications in hydrology and hydrogeology, for example groundwater flows through sediments that need to be studied under various geometric scenarios.}

Here, our goal is to learn the operator of the system in Eq.~\eqref{eq:Darcy}, which maps the input random conductivity field to the output hydraulic head, i.e., $\mathcal{G}_\theta: K(\boldsymbol x) \rightarrow h(\boldsymbol x)$. To generate multiple conductivity fields to train the DeepONet, we describe $K(\boldsymbol x)$ as a stochastic process, \review{the realizations of which are generated with a truncated Karhunen-Lo\'eve expansion (KLE)}. The source simulation box is a square domain $\Omega= [0,1] \times [0,1]$, discretized with $d=1541$ grid points. We consider the following four transfer learning scenarios, which are also visually presented in Figure \ref{fig:applications}:
\begin{itemize}
    \item \textbf{TL1:} Transfer learning from a square domain to an equilateral triangle. 
    \item \textbf{TL2:} Transfer learning from a square domain to a right-angled triangle.
    \review{\item \textbf{TL3:} Transfer learning from a square domain to an equilateral triangle with a vertical notch.}
    \item \textbf{TL4:} Transfer learning from a square domain with one vertical notch to a square domain with two horizontal notches.
\end{itemize}
\begin{table}[ht!]
\footnotesize
\begin{center}
\begin{minipage}{\textwidth}
\caption{\Review{Relative $L_2$ error (\%) for all Darcy flow problems (TL1 - TL4).}}
\begin{tabular}{c c c c c c}
\toprule
& $N_t$  & TL1 & TL2 & \review{TL3} & TL4 \\   
 \toprule
\begin{tabular}{@{}c@{}}Training DeepONet \\ (source) \end{tabular} & $2{\small,}000$  &  $1.36 \pm 0.48$ &  $1.36 \pm 0.48$  &  $1.36 \pm 0.48$ & $1.58 \pm 0.05$ \\ \hdashline
\multirow{7}{*}{\begin{tabular}{@{}c@{}}Training DeepONet \\ (target)\end{tabular}} 
& $5$    & $31.60 \pm 0.36$ & $25.53 \pm 1.41$ &  $29.14 \pm 0.52$ & $51.70 \pm 0.18$ \\ 
& $20$   & $26.50 \pm 0.62$  & $19.68 \pm 0.39$ & $25.57 \pm 0.43$ & $39.69 \pm 0.13$ \\ 
& $50$   & $26.41 \pm 0.54$ & $16.68 \pm 0.34$  & $18.68 \pm 0.36$ & $25.82 \pm 0.62$ \\
& $100$  & $19.20 \pm 0.36$  & $15.39 \pm 0.26$  & $15.56 \pm 0.34$ & $19.50 \pm 0.25$ \\
& $150$  & $14.40 \pm 0.33$ & $13.20 \pm 0.22$  & $14.71 \pm 0.30$ & $16.82 \pm 0.23$ \\
& $200$  & $11.90 \pm 0.32$  & $12.03 \pm 0.09$ & $12.32 \pm 0.29$ & $12.00 \pm 0.27$ \\
& $250$  & $9.50 \pm 0.10$ & $8.32 \pm 0.09$  & $11.75 \pm 0.22$ & $9.22 \pm 0.04$ \\
& $2{\small,}000$  &  $1.40 \pm 0.10$ & $1.62 \pm 0.30$  &  $1.83 \pm 0.37$ & $1.78 \pm 0.18$  \\\hdashline
\multirow{7}{*}{\begin{tabular}{@{}c@{}}Training \\ TL-DeepONet\end{tabular}} & $5$  &  $10.70 \pm 0.53$ & $11.73 \pm 1.59$ &  $12.28 \pm 0.44$ & $10.46 \pm 0.93$ \\ 
& $20$ & $8.81 \pm 0.23$  & $8.79 \pm 0.30$ & $9.54 \pm 0.05$ & $8.16 \pm 0.29$ \\ 
& $50$  & $7.37 \pm 0.18$ & $7.42 \pm 0.48$  & $7.54 \pm 0.76$ & $7.01 \pm 0.17$ \\
& $100$  & $6.51 \pm 0.28$  & $6.51 \pm 0.28$  & $6.45 \pm 0.36$ & $5.78 \pm 0.42$ \\
& $150$  & $5.61 \pm 0.30$ & $5.54 \pm 0.33$  & $6.04 \pm 0.28$ & $5.10 \pm 0.15$ \\
& $200$  & $3.85 \pm 0.07$  & $4.44 \pm 0.32$ & $5.28 \pm 0.19$ & $4.11 \pm 0.26$ \\
& $250$  & $3.74 \pm 0.07$ & $3.46 \pm 0.27$  & $4.92 \pm 0.15$ & $3.34 \pm 0.09$ \\
& \review{ $2{\small,}000$}  & \review{$3.19 \pm 0.14$}  & \review{$2.19 \pm 0.17$} & \review{$4.72 \pm 0.22$} & \review{$2.14 \pm 0.03$} \\
\bottomrule
\end{tabular}
\label{table:TL1-4}
\end{minipage}
\end{center}
\end{table}

\begin{figure}[ht!]
\begin{center}
\includegraphics[width=0.7\textwidth]{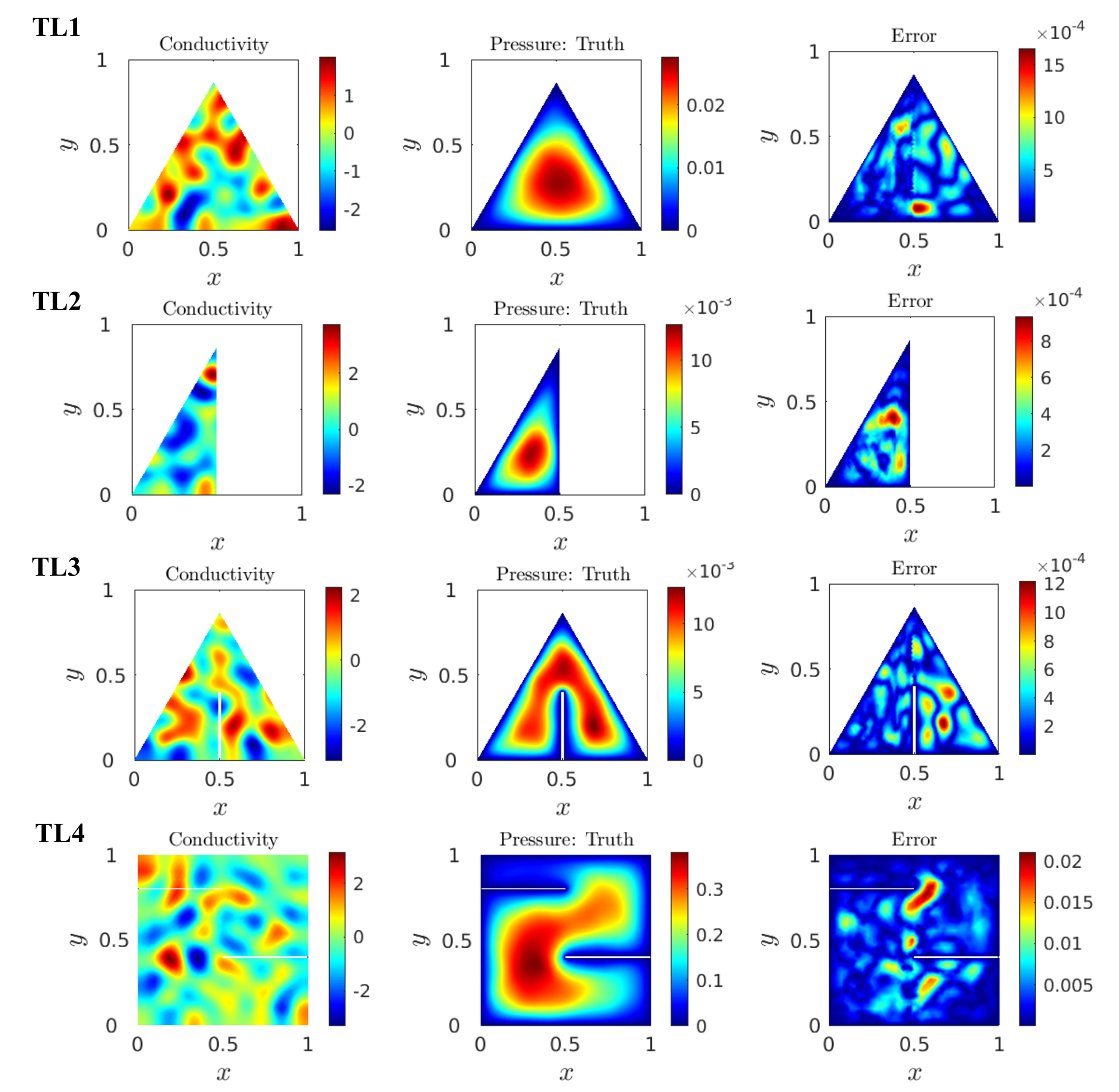}
\caption{Representative results for the Darcy model (TL1-TL4) takes as input the spatially varying conductivity field and approximates the hydraulic head over the domain. \review{Error fields represent the point-wise error computed as $\lvert \frac{f(\mathbf{x}^{T})- {\mathbf{y}^T}}{{\mathbf{y}^T}}\rvert $, where $\mathbf{y}^T$, $f(\mathbf{x}^{T})$ is the reference response and the model prediction respectively.}}
\label{fig:results-TL1-4}
\end{center}
\end{figure}

\begin{table}[ht!]
\footnotesize
\centering
\caption{Training cost in seconds ($s$) for all Darcy flow problems (TL1 - TL4).}
\begin{tabular}{c c c c c c}
\toprule
& $N_t$  & TL1 & TL2 & \review{TL3} & TL4 \\   
 \toprule
\begin{tabular}{@{}c@{}}Training DeepONet \\ (source) \end{tabular} & $2{\small,}000$  &  $15{\small,}260$ &  $15{\small,}260$  &  $15{\small,}260$ & $2{\small,}261$  \\ \hdashline
\begin{tabular}{@{}c@{}}Training DeepONet \\ (target)\end{tabular}
& $2{\small,}000$  &  $12{\small,}880$ & $18{\small,}200$  &  $18{\small,}080$ & $3{\small,}978$  \\ \hdashline
\multirow{7}{*}{\begin{tabular}{@{}c@{}}Training \\ TL-DeepONet\end{tabular}} & $5$  &  $11$ & $10$ &  $10$ & $83$ \\ 
& $20$ & $129$  & $116$ & $112$ & $139$ \\ 
& $50$  & $416$ & $399$  & $302$ & $289$ \\
& $100$  & $439$  & $437$  & $351$ & $300$ \\
& $150$  & $459$ & $439$  & $378$ & $302$ \\
& $200$  & $462$  & $480$ & $406$ & $304$ \\
& $250$  & $531$ & $528$  & $586$ & $305$ \\
& \review{ $2{\small,}000$}  & \review{$595$}  & \review{$601$} & \review{$653$} & \review{$350$} \\
\bottomrule
\end{tabular}
\label{table:TL1-4-time}
\end{table}
To train the source model, DeepONet, we \review{sample realizations of the random conductivity field $K(\boldsymbol x)$ and corresponding model response $h(\boldsymbol x)$ to} generate $N_s=2{\small,}000$ source data and test on an additional set of $N^s_{\text{test}}=100$ samples. Furthermore, we generate $N_t=2{\small,}000$ target data and $N^t_{\text{test}}=100$ additional target test data. \review{The realizations of the conductivity field for the target domain are obtained by a linear interpolation of the structured source domain to an unstructured target domain.} For all PDE benchmarks, \Review{DeepONet and} TL-DeepONet, \Review{are} trained for $N_t=\{5,20,50,100,150,200,250,2{\small,}000\}$ samples to evaluate the effect of the size of the train dataset. For all applications, we employ a \review{convolutional neural network (CNN)} for the branch net and a \review{feedforward neural network (FNN)} for the trunk net of the operator network. We train all models on a single NVIDIA RTX A6000 GPU until convergence of the loss function and provide the wall clock time in seconds.
\begin{table}[ht!]
\footnotesize
\begin{center}
\caption{\review{Relative $L_2$ error (\%) and training cost in seconds (s) for training the target domain without $\mathcal L_{\text{CEOD}}$ for the Darcy problem on a triangular domain with a notch (TL3).}}
\begin{tabular}{c c c c}
\toprule
& $N_t$  & $L_2$ ($\%$) (without $\mathcal L_{\text{CEOD}}$) & time (s) \\   
 \toprule
\multirow{7}{*}{\begin{tabular}{@{}c@{}}Training \\ TL-DeepONet \end{tabular}} & $5$  &  $66.27 \pm 0.53$ & $13$\\ 
& $20$ & $62.10 \pm 0.29$ & $91$\\ 
& $50$  & $55.51 \pm 0.29$ & $156$\\
& $100$  & $48.85 \pm 0.28$ & $202$\\
& $150$  & $41.61 \pm 0.23$ & $276$\\
& $200$  & $33.85 \pm 0.21$ & $301$\\
& $250$  & $26.45 \pm 0.17$ & $345$\\
&  $2{\small,}000$  & $14.29 \pm 0.14$ & $445$ \\
\bottomrule
\end{tabular}
\label{table:TL3_withoutCEOD}
\end{center}
\end{table}
\review{The relative $L_2$ norm error ($\%$) and training cost ($s$) for all TL scenarios are presented in Tables \ref{table:TL1-4} and \ref{table:TL1-4-time} respectively.
The reported errors represent the mean $\pm$ one standard deviation based on five runs with different seed numbers to account for the stochasticity of the training process. \Review{We observe that for a small number of training samples, training DeepONet from scratch consistently results in overfitting of the model while  TL-DeepONet performs significantly better.} Training DeepONet on the target domain with random initialization and $N_t=2{\small,}000$ training samples results in $1.4\%$, $1.62\%$, $1.83\%$ and $1.87\%$ mean relative $L_2$ error for the four TL tasks, respectively and a high computational cost. By utilizing the learned source operator by transferring the trained parameters to the target model leads to a very good accuracy even when only scarce data are available. We observe that with $N_t = 250$ samples, the proposed transfer learning framework allows us to achieve small errors (relative $L_2<5\%$).} When TL-DeepONet is trained with $N_t=2{\small,}000$ training samples we observe that for these TL scenarios, the model achieves very good accuracy. As expected, the optimal accuracy achieved via transfer learning is lower than the accuracy of training the target model from scratch for equal number of training data \review{given the constrained expressivity of the TL-DeepONet}. However, this discrepancy is rather small. \review{From Table \ref{table:TL1-4-time}, we observe that training TL-DeepONet through TL is much faster than training from scratch, even when $2{\small,}000$ training data are considered. This is also expected, as training a smaller FNN results in less training cost and in addition, training of the original source DeepONet is initially required. However, when considering multiple target tasks, TL-DeepONet can simultaneously achieve very good performance with a significantly smaller overall training cost. Finally, in Table \ref{table:TL3_withoutCEOD} the results of training TL-DeepONet without the addition of $\mathcal L_{\text{CEOD}}$ are shown. We observe that the model performs significantly worse when trained with a simple regression loss, for all tested number of training data ($N_t$).} Representative results for each task-specific learning are shown in Figure \ref{fig:results-TL1-4}. 

\bigbreak
\noindent
\textbf{Elasticity model} 

\noindent
We consider a thin rectangular plate subjected to in-plane loading that is modeled as a two-dimensional problem of plane stress elasticity. The relevant equations are given below:
\label{subsec:elasticity}
\begin{equation}
\label{eq:elasticity}
    \review{\nabla \cdot \boldsymbol{\sigma} + \boldsymbol{f}(\boldsymbol x) = 0},\hspace{10pt} \boldsymbol x = (x,y),
\end{equation}
\begin{equation*}
\label{eq:elasticity-BC}
    \review{(u, v) = 0}, \hspace{10pt} \forall\;\; x=0,
\end{equation*}
where \review{$\boldsymbol{\sigma}$ is the Cauchy stress tensor, $\boldsymbol{f}$ is the body force,} $u(\boldsymbol x)$ and $v(\boldsymbol x)$ represent the $x$- and $y$-displacement, respectively. In addition, $E$, and $\nu$ represent the Young modulus and Poisson ratio of the material, respectively. The relation among stress and displacement in plane stress condition is defined as:
\review{
\begin{equation}
    \begin{Bmatrix}
        \sigma_{xx}\\ \sigma_{yy}\\ \tau_{xy}
    \end{Bmatrix} = \frac{E}{1-\nu^2}
    \begin{bmatrix}
        1 & \nu & 0 \\ \nu & 1 & 0 \\ 0 & 0 & \frac{1-\nu}{2}
    \end{bmatrix} \times     
    \begin{Bmatrix}
        \frac{\partial u}{\partial x} \\ \frac{\partial v}{\partial y} \\ \frac{\partial u}{\partial y} + \frac{\partial v}{\partial x}
    \end{Bmatrix}.
\end{equation}} 

For the source model, we consider a plate with a centered circular internal boundary in the simulation box subjected to uniaxial tension. The presence of a circular hole in the plate in the source domain disrupts the uniform stress distribution near the hole resulting in significantly higher than average stress. \review{In practice, transfer learning can be useful in mechanics applications such as this for design optimization to predict mechanical deformation in objects under multiple design parameters, including geometry and material properties.}
\begin{table}[ht!]
\footnotesize
\caption{\Review{Relative $L_2$ error and training cost in seconds ($s$) for the elasticity transfer learning problem (TL5).}}
\centering
\begin{tabular}{c c c c c}
\toprule
& \multirow{2}{*}{$N_t$}  & \multicolumn{3}{c}{$L_2$ ($\%$)} \\  \cmidrule(l){3-5} & & $u(\boldsymbol x)$ & $v(\boldsymbol x)$ & time ($s$)  \\
 \toprule
\begin{tabular}{@{}c@{}}Training DeepONet \\ (source)\end{tabular} & $1{\small,}900$  &  $2.26 \pm 0.09$ & $2.52 \pm 0.18$ & $10{\small,}884$ \\ \hdashline
\multirow{7}{*}{\begin{tabular}{@{}c@{}}Training DeepONet\\ (target)\end{tabular}} 
& $5$  &  $198.36 \pm 34.4$ & $296.0 \pm 45.90$ & $5{\small,}980$  \\  
& $20$ & $67.70 \pm 12.30$ & $91.68 \pm 30.81$ & $6{\small,}517$ \\  
& $50$  & $21.90 \pm 3.36$ & $24.3 \pm 5.16$ & $6{\small,}900$ \\
& $100$  & $11.6 \pm 3.59$ & $13.26 \pm 3.07$ & $7{\small,}168$  \\
& $150$  & $8.42 \pm 0.81$ & $12.34 \pm 3.41$ & $7{\small,}364$ \\
& $200$  & $7.34 \pm 0.05$ & $9.18 \pm 1.80$ & $8{\small,}924$ \\
& $250$  & $5.98 \pm 0.05$ & $7.16 \pm 0.86$ & $8{\small,}997$ \\
& $1{\small,}900$  &  $1.72 \pm 0.08$ & $2.80 \pm 0.25$ & $10{\small,}780$ \\ \hdashline
\multirow{7}{*}{\begin{tabular}{@{}c@{}}Training \\ TL-DeepONet\end{tabular}} & $5$  &  $8.17 \pm 0.05$ & $11.31 \pm 1.36$ & 53  \\  
& $20$ & $5.94 \pm 0.33$ & $10.7 \pm 0.27$ & $148$ \\  
& $50$  & $4.72 \pm 0.44$ & $9.84 \pm 0.05$ & $264$ \\
& $100$  & $4.42 \pm 0.24$ & $8.58 \pm 0.15$ & $376$  \\
& $150$  & $4.14 \pm 0.05$ & $7.2 \pm 0.01$ & $418$ \\
& $200$  & $3.84 \pm 0.05$ & $6.11 \pm 0.02$ & $509$ \\
& $250$  & $3.56 \pm 0.05$ & $4.33 \pm 0.05$ & $515$ \\
& \review{$1{\small,}900$ } & \review{$2.50 \pm 0.01$} & \review{$2.84 \pm 0.05$} &  \review{$553$} \\
\bottomrule
\end{tabular}
\label{table:TL5}
\end{table}
\begin{table}[ht!]
\small
\caption{\Review{Relative $L_2$ error and training cost in seconds ($s$) for the elasticity transfer learning problem (TL6).}}
\centering
\footnotesize
\begin{tabular}{c c c c c}
\toprule
& \multirow{2}{*}{$N_t$}  & \multicolumn{3}{c}{$L_2$ ($\%$)} \\  \cmidrule(l){3-5} & & $u(\boldsymbol x)$ & $v(\boldsymbol x)$ & time ($s$)  \\
 \toprule
\begin{tabular}{@{}c@{}}Training DeepONet \\ (source)\end{tabular} & $1{\small,}900$  &  $2.30 \pm 0.49$ & $3.22 \pm 0.48$ & $10{\small,}060$ \\ \hdashline
\multirow{7}{*}{\begin{tabular}{@{}c@{}}Training DeepONet \\ (target)\end{tabular}} 
& $5$  &  $137.5 \pm 31.15$ & $171.9 \pm 24.07$ & $9{\small,}570$  \\  
& $20$ & $131.36 \pm 83.07$ & $149.66 \pm 54.49$ & $9{\small,}368$ \\  
& $50$  & $28.14 \pm 8.3$ & $27.8 \pm 7.43$ & $9{\small,}487$ \\
& $100$  & $10.62 \pm 1.64$ & $11.44 \pm 1.38$ & $9{\small,}500$  \\
& $150$  & $8.76 \pm 1.67$ & $11.02 \pm 1.38$ & $10{\small,}534$ \\
& $250$  & $9.14 \pm 0.60$ & $10.7 \pm 1.41$ & $10{\small,}583$ \\
& $1{\small,}900$  &  $2.72 \pm 0.26$ & $1.92 \pm 0.41$ & $11{\small,}750$ \\ \hdashline
\multirow{7}{*}{\begin{tabular}{@{}c@{}}Training \\ TL-DeepONet\end{tabular}} & $5$  &  $281.9 \pm 1.15$ & $178.2 \pm 1.36$ & $53$  \\  
& $20$ & $17.30 \pm 0.19$ & $23.90 \pm 0.17$ & $148$ \\  
& $50$  & $16.01 \pm 0.12$ & $15.11 \pm 0.15$ & $264$ \\
& $100$  & $15.04 \pm 0.13$ & $12.50 \pm 0.15$ & $376$  \\
& $150$  & $13.14 \pm 0.05$ & $11.97 \pm 0.09$ & $418$ \\
& $250$  & $11.12 \pm 0.04$ & $9.66 \pm 0.14$ & $480$ \\
& $1{\small,}900$ & $7.62 \pm 0.04$ & $6.1 \pm 0.10$ &  $512$ \\
\bottomrule
\end{tabular}
\label{table:TL6}
\end{table}

We model the loading conditions $f(\boldsymbol x)$ applied to the right edge of the plate as a Gaussian random field (see Figure \ref{fig:applications}). We aim to learn the mapping from the random boundary load to the displacement field ($\boldsymbol u$: $x$-displacement and $\boldsymbol v$: $y$-displacement), such that $\mathcal G_{\theta}:f(\boldsymbol x)\rightarrow [ \boldsymbol u(\boldsymbol x), \boldsymbol v(\boldsymbol x)]$. Therefore, we train the DeepONet surrogate to predict two distinct model outputs. In this example, we consider the following two TL scenarios:
\begin{itemize}
    \item \textbf{TL5:} Transfer learning from a domain with a centered circular internal boundary and material properties \review{$(E_S=300\times10^5,\nu_S=0.3)$} to a domain with two smaller circular internal boundaries in the upper right and lower left corners and different material properties \review{$(E_T=410\times10^3,\nu_T=0.35)$}.
    \item \review{\textbf{TL6:} Transfer learning from a domain with a centered circular internal boundary and material properties \review{$(E_S=300\times10^5,\nu_S=0.3)$} to a domain with a square internal boundary and different material properties \review{$(E_T=410\times10^3,\nu_T=0.45)$}.}
\end{itemize}
\review{Compared to the previous application, here the TL scenario is more challenging as source and target domains differ not only in the (internal) boundary geometry but also in the model parameters.} To train the source model, we generate $N_s=1{\small,}900$ source data and test on an additional set of $N^s_{\text{test}}=100$ samples. Furthermore, we generate $N_t=1{\small,}900$ target data and $N^t_{\text{test}}=100$ additional test data. Detailed results are shown in Tables \ref{table:TL5}, and \ref{table:TL6}. We note that in the DeepONet's branch net, before the input to the CNN, a linear layer is used to project the boundary to a square domain. For TL5, transfer learning allows accurate predictions of the displacement fields $u(\boldsymbol x),v(\boldsymbol x)$ under the case of scarce data (e.g., $250$ target samples). \review{In Table \ref{table:TL5}, we observe that for $N_t=1\small,900$ the resulting TL-DeepONet accuracy is similar to the case of training DeepONet on the target domain, however transfer learning results in $\sim 20\times$ training speedup. These findings become all the more important especially in cases where one aims to learn multiple tasks, as the proposed framework removes the need of training models from scratch for each target task.} 

\begin{figure}[ht!]
\begin{center}
\includegraphics[width=0.85\textwidth]{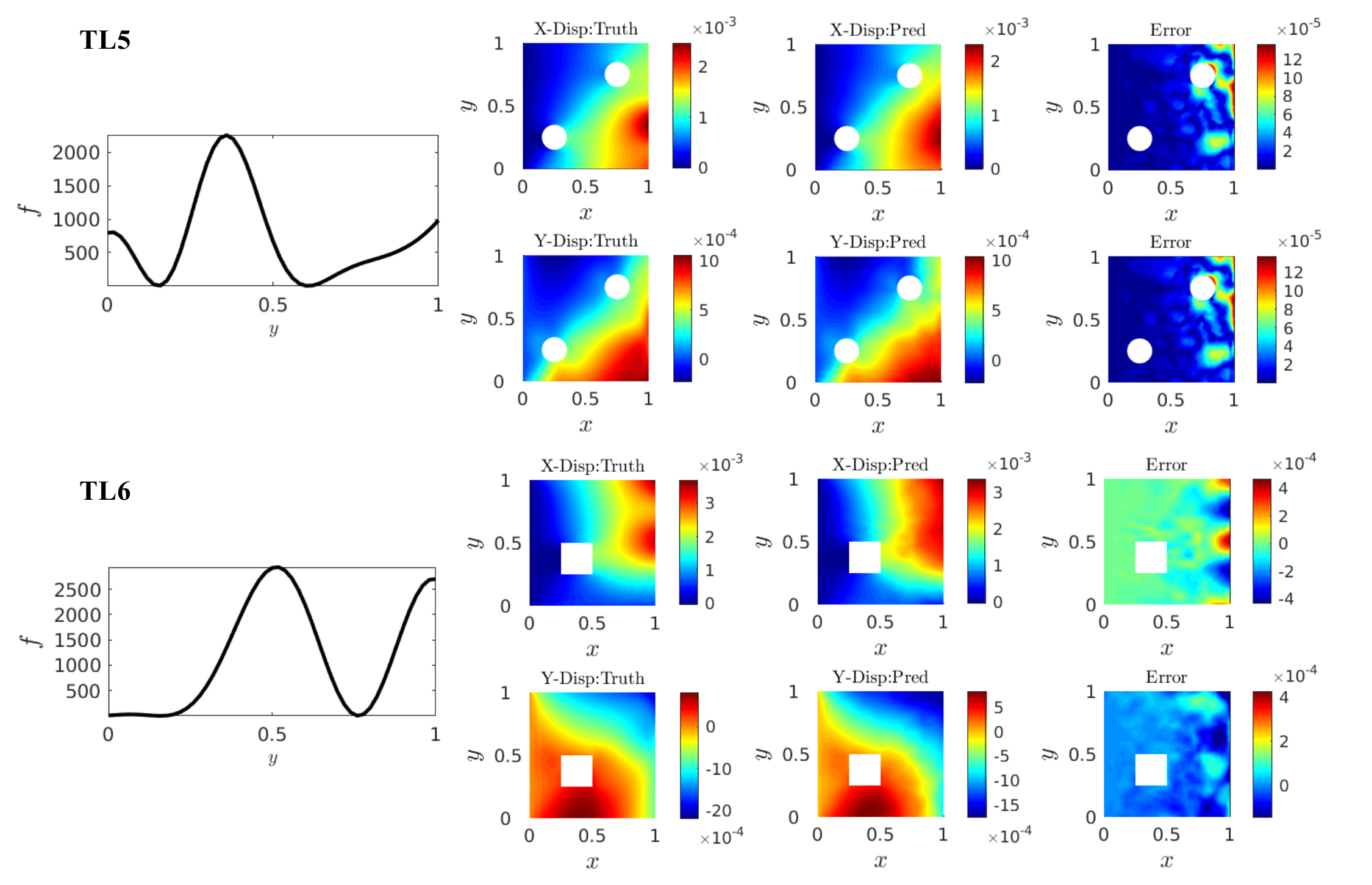}
\caption{\review{Representative results for the elasticity model (TL5 and TL6) for which DeepONet takes as input the loading condition applied on the right edge of the plate (left frames) and outputs the displacement field (middle frames). \review{Error fields, shown in the right frame, represent the point-wise error computed as $({\mathbf{y}^T} - f(\mathbf{x}^{T})) $, where $\mathbf{y}^T$, $f(\mathbf{x}^{T})$ is the reference response and the model prediction respectively.}.}}
\label{fig:results-TL5-6}
\end{center}
\end{figure}

\review{For task TL6, the main challenge stems from the significant change both in the shape of the internal boundaries and the material properties ($\nu_S=0.3$ and $\nu_T=0.45$). TL-DeepONet results in a relative $L_2$ error of $7.62 \pm 0.04$ and $6.1 \pm 0.10$ for the horizontal and vertical displacement respectively and for $1\small,900$ training data (see Table \ref{table:TL6}). We observe that when the model parameters are changed to a large extent the predictive accuracy of TL-DeepONet deteriorates. In such cases, transferring domain-invariant features extracted from the source model is insufficient to allow the TL-DeepONet model to result in small error after finetuning. This demonstrates one of the limitations of the proposed transfer learning approach.} Representative results for a given input realization are shown in Figure \ref{fig:results-TL5-6}. We observe that the proposed TL approach allows for task-specific learning even in cases where the source and target domains are characterized by different internal boundaries and material properties, \review{with some limitations. This example is extended in TL12 of the SI, where we take into account changes to both geometry and material characteristics, but the changes to material properties (Poisson's ratio) are only made within certain bounds. The target domain has $\nu_T=0.35$ while the source domain has $\nu_S=0.3$. With this extension we illustrate that TL-DeepONet is efficient in capturing related geometrical changes and changes in the material properties within a range.}

\bigbreak
\noindent
\textbf{Brusselator diffusion-reaction system} 

\noindent
Finally, we consider the Brusselator diffusion-reaction system, which describes an autocatalytic chemical reaction in which a reactant substance interacts with another substance to increase its production rate \cite{ahmed2019numerical}. The Brusselator system is characterized by the following reactions:
\begin{subequations}\label{eq:reaction_model}
\begin{align}
    A & \xrightarrow{k_1} X, \quad \\ 
    B + X  & \xrightarrow{k_2} Y + D, \\
    2X + Y & \xrightarrow{k_3} 3X,\\
    X & \xrightarrow{k_4} E,
\end{align}
\label{eq:A3-reaction}
\end{subequations}
where $k_i, (i= 1, 2, 3, 4)$ are positive parameters representing the reaction rate constant. In Eq.~\eqref{eq:reaction_model}, a reactant, $A$, is converted to a final product $E$ in four steps with the help of four additional species $X,\;B,\;Y,$ and $D$. Species $A$ and $B$ are in vast excess and thus can be modeled at constant concentration. The two-dimensional rate equations read:
\begin{equation}
\begin{split}
    \frac{\partial u}{\partial t} &= D_0 \bigg(\frac{\partial^2 u}{\partial x^2} + \frac{\partial^2 u}{\partial y^2} \bigg) + a - (1+b)u + vu^2, \\
    \frac{\partial v}{\partial t} &= D_1 \bigg(\frac{\partial^2 v}{\partial x^2} + \frac{\partial^2 v}{\partial y^2} \bigg) + bu - vu^2, \hspace{10pt} \boldsymbol x \in [0,1]^2, \ t \in [0,1],
\end{split} 
\label{eq:A3-model}
\end{equation}
subject to the following initial conditions:
\begin{align*}
u(\boldsymbol x,t=0) &= h_1(\boldsymbol x) \ge 0, \\
v(\boldsymbol x,t=0) &= h_2(\boldsymbol x) \ge 0,
\end{align*}
where $\boldsymbol x=(x,y)$ are the spatial coordinates, $D_0, D_1$ represent the diffusion coefficients, $a=\{A\}, b=\{B\}$ are constant concentrations, and $u=\{X\}, v=\{Y\}$ represent the concentrations of reactant species $X,Y$. \review{In process systems engineering where the goal is to design, control and optimize chemical physical and biological processes described by dynamical systems, transfer learning can provide a useful means of learning the system dynamics under various scenarios (e.g., different number of species, thermodynamic properties etc.).}

\begin{figure}[ht!]
\begin{center}
\includegraphics[width=0.8\textwidth]{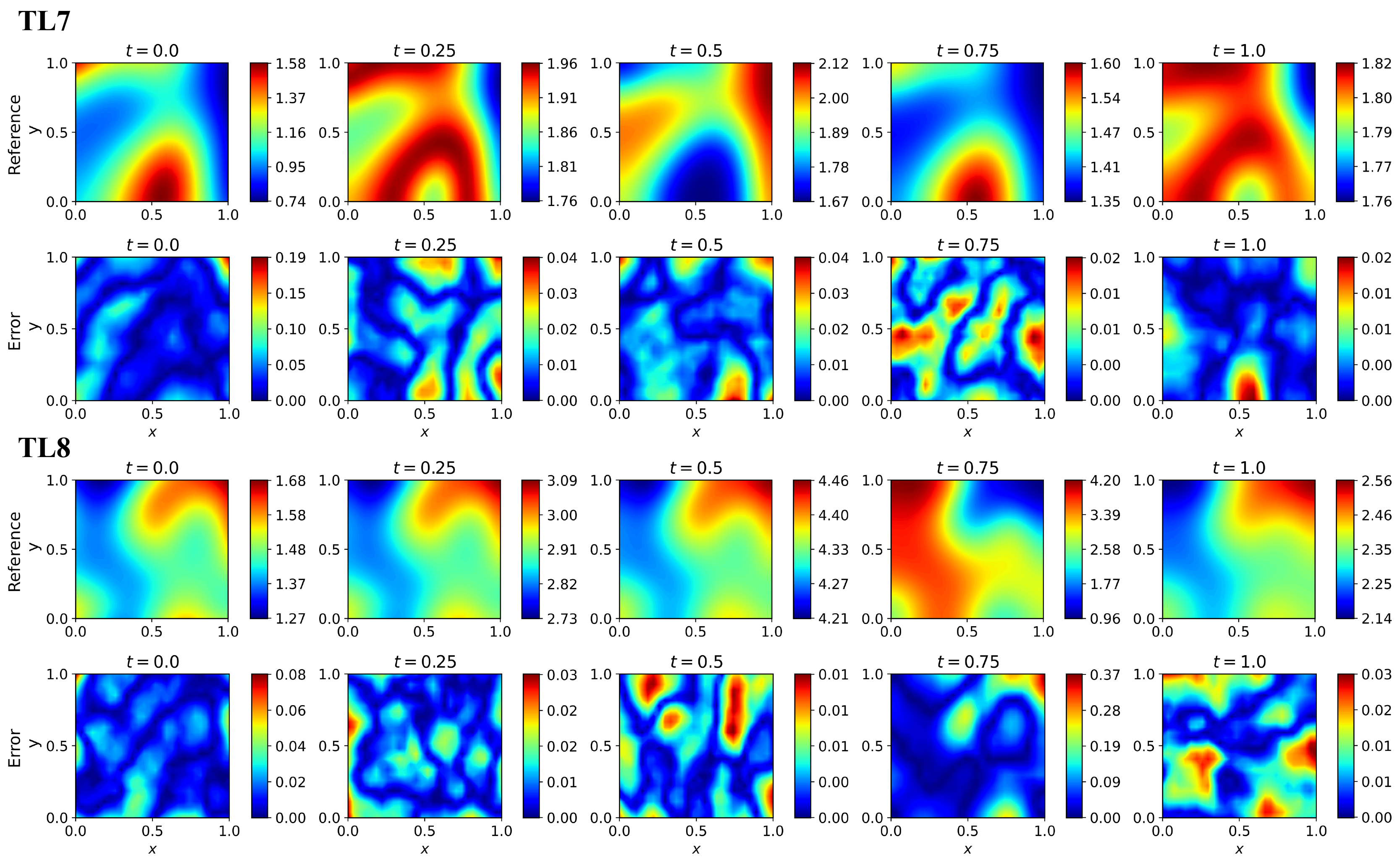}
\caption{\review{Representative result for the Brusselator reaction-diffusion system transfer learning problem TL7 and TL8 showing a reference realization and the TL-DeepONet error. This problem takes as input the initial random field depicting the concentration of one of the species. TL7 approximates the transfer of knowledge from a system with damped oscillations to overdamped oscillations, whereas TL8 represents the transfer to a system with periodic oscillations}. \review{Error fields represent the point-wise error computed as $\lvert \frac{f(\mathbf{x}^{T})- {\mathbf{y}^T}}{{\mathbf{y}^T}}\rvert $, where $\mathbf{y}^T$, $f(\mathbf{x}^{T})$ is the reference response and the model prediction, respectively.}}
\label{fig:results-TL7-8}
\end{center}
\end{figure}

In this problem, we train DeepONet to learn the mapping between the initial field and the evolved concentration of species $v$, i.e.\ $\mathcal G_{\theta}: h_2(\boldsymbol x) \rightarrow v(\boldsymbol x, t)$. The initial field $h_2(x,y)$ is modeled as a Gaussian random field. The following two transfer learning problems are considered (see Figure \ref{fig:applications}):
\begin{itemize}
    \item \textbf{TL7:} Transfer learning from damped oscillations to overdamped oscillations  (approaching fast a steady-state response). 
    \item \textbf{TL8:} Transfer learning from damped oscillations to periodic oscillations (limit cycle in phase space).
\end{itemize}
The Brusselator dynamics can be controlled by the constant concentration $b=\{B\}$, which we set as $b_S=2.2$ and $b_{T_1}=1.7, b_{T_2}=3.0$ for the source and two target tasks respectively. For the source model, we generate $N_s=800$ training and $N^s_{\text{test}}=200$ test data. In addition, we generate $N_t=800$, $N^t_{\text{test}}=200$ training and test target data for training the target model. The results for both TL5 and TL6 problems are presented in Table \ref{table:TL5-6}.
\begin{table}[ht!]
\footnotesize
\caption{\Review{Relative $L_2$ error and training cost in seconds ($s$) for Brusselator transfer learning problems (TL7 \& TL8).}}
\centering
\begin{tabular}{c c c c c c}
\toprule
& \multirow{2}{*}{\begin{tabular}{@{}c@{}}\# of training \\ data ($N_t$)\end{tabular}}  & \multicolumn{2}{c}{TL7} & \multicolumn{2}{c}{TL8} \\  \cmidrule(l){3-4}  \cmidrule(l){5-6}  & & $L_2$ ($\%$)  & time ($s$) & $L_2$ ($\%$) & time ($s$)  \\
 \toprule
\begin{tabular}{@{}c@{}}Training DeepONet \\ (source)\end{tabular} & $800$  &  $1.51 \pm 0.11$ & $3{\small,}532$ & $1.51 \pm 0.11$  & $3{\small,}532$ \\ \hdashline
\multirow{7}{*}{\begin{tabular}{@{}c@{}}Training DeepONet \\ (target)\end{tabular}}
& $5$  &  $71.91 \pm 0.58$ & $1279$ & $73.87 \pm 0.92$ & $2415$  \\  
& $20$ &  $31.85 \pm 0.32$ & $1263$ & $52.79 \pm 0.38$ & $2425$  \\  
& $50$  & $26.22 \pm 0.42$ & $1251$ & $36.77 \pm 0.37$ & $2527$  \\
& $100$  & $19.64 \pm 0.36$ & $1242$ & $26.22 \pm 0.37$ & $2705$  \\
& $150$  & $16.82 \pm 0.23$ & $1231$ & $19.50 \pm 0.25$ & $2837$  \\
& $200$  & $10.6 \pm 0.16$ & $1207$ & $14.84 \pm 0.30$ & $2987$  \\
& $250$  & $7.46 \pm 0.10$ & $1201$ & $10.56 \pm 0.34$ & $3019$ \\
& $800$  &  $1.16 \pm 0.05$ & $2{\small,}461$ & $2.92 \pm 0.04$  & $4{\small,}754$  \\ \hdashline
\multirow{7}{*}{\begin{tabular}{@{}c@{}}Training \\ TL-DeepONet \end{tabular}} 
& $5$  &  $21.66 \pm 1.54$ & $57$ & $48.7 \pm 5.48$ & $81$  \\  
& $20$ & $13.54 \pm 1.35$ & $132$ & $16.36 \pm 1.31$ & $170$  \\  
& $50$  & $2.41 \pm 0.02$ & $188$ & $13.04 \pm 1.15$ & $190$  \\
& $100$  & $2.32 \pm 0.02$ & $206$ & $9.60 \pm 0.38$ & $212$  \\
& $150$  & $2.26 \pm 0.08$ & $209$ & $8.96 \pm 0.88$ & $288$  \\
& $200$  & $2.12 \pm 0.04$ & $213$ & $5.06 \pm 0.13$ & $310$  \\
& $250$  & $2.02 \pm 0.04$ & $229$ & $3.78 \pm 0.16$ & $345$ \\
& \review{$800$} & \review{$1.98 \pm 0.01$} & \review{$260$}  & \review{$3.39 \pm 0.15$} & \review{$532$}  \\
\bottomrule
\end{tabular}
\label{table:TL5-6}
\end{table}
For the case where the TL task involves learning the overdamped oscillations (TL7), we observe that TL-DeepONet achieves very high accuracy for $N_t\ge50$ samples. Interestingly, as we increase the number of target training samples, the relative $L_2$ error converges rapidly to the result with random initialization and a plethora of available data (see Table \ref{table:TL5-6}, second row). In contrast, for the more challenging task (TL8), where TL-DeepONet learns the oscillatory response, more available data are needed for accurate operator regression. \review{All surrogate models are also tested on two out-of-distribution (OOD) datasets to evaluate the ability of models to extrapolate. These results are shown in the Supplementary Information Table \ref{table:OOD}.} Representative results for both TL scenarios are presented in Figure \ref{fig:results-TL7-8}, where the reference and point-wise error fields are shown for five distinct time steps.

\begin{table}[ht!]
\footnotesize
\caption{\review{Uncertainty propagation and moment estimation of the Brusselator response by comparing standard MCS with DeepONet trained on target domain and TL-DeepONet.}}
\centering
\begin{tabular}{c c c c c c c}
\toprule
  & \multicolumn{4}{c}{Uncertainty propagation ($t=0.5$)} & \multicolumn{2}{c}{Moment estimation} \\  \cmidrule(l){2-5}  \cmidrule(l){6-7} &
  \multicolumn{2}{c}{$\boldsymbol x_1 =(0.37,0.741)$}  & \multicolumn{2}{c}{$\boldsymbol x_2 =(0.926,0.296)$} & \multicolumn{2}{c}{$L_2$ error w/ MCS} \\
  & $\mu$ &  $\sigma^2$ & $\mu$ & $\sigma^2$ & $\mu$ (field) & $\sigma^2$ (field) \\
 \toprule
\begin{tabular}{@{}c@{}}Monte Carlo \\ simulation (MCS)\end{tabular}  &  $4.339$ & $0.038$ & $4.345$  & $0.041$ & -- & -- \\ \hdashline
\begin{tabular}{@{}c@{}}Training DeepONet \\ (target)\end{tabular}   & $4.328$ &  $0.040$ & $4.327$ & $0.038$ & $0.015$  & $0.125$  \\ \hdashline
\begin{tabular}{@{}c@{}}Training  \\ TL-DeepONet \end{tabular}  & $4.330$ & $0.040$ & $4.340$ & $0.035$ & $0.030$ & $0.257$  \\  
\bottomrule
\end{tabular}
\label{table:UQ}
\end{table}
Furthermore, in this example, we also compare the results with a conventional transfer learning approach proposed in \cite{goswami2020transfer}. In this work, the target domain has the same geometrical definition as the source domain. While no additional loss term is presented, it instead proposes to train the target domain with the same network as the source domain, while freezing the network parameters of all the layers except the last layer. To compare the approach proposed in \cite{goswami2020transfer} with TL-DeepONet, we trained the target model by retraining the weights and biases associated with the last layer only. The weights and biases corresponding to the other layers of the target model are kept fixed at previously trained values obtained from the source model. Implementing this approach results in a relative $L_2$ error of $7.45\%$ and $10.92\%$ for TL7 and TL8, respectively. Thus, incorporating the discrepancy loss as a part of the total loss, results in a better accuracy for transfer learning tasks in operator regression problems involving PDEs.

\review{With TL8, we also demonstrate the efficiency of TL-DeepONet to take into account the effect of non-linearities in the target domain, which considers the parameter $b = 3.0$. The obtained results prove that the proposed approach is efficient in transferring knowledge to target domains. We have also added the Burgers equation (TL13) in the SI to consider different types of non-linearity. Our objective in TL13 is to map the random input fields to the model solution at the final simulation time step. While the source model data are generated for a high diffusion coefficient ($\nu=0.2$), for the target data we are considering a much smaller value ($\nu=0.001/\pi$), which results in a highly non-linear response. The results demonstrate that TL-DeepONet is efficient in handling non-linearities induced in the target domain.}

\review{Finally, in context of UQ, we aim to perform uncertainty propagation and moment estimation to capture the statistics of the quantities of interest (QoI), $v(\boldsymbol x,t)$. In Table \ref{table:UQ}, we compare the results of standard Monte Carlo simulation (MCS) from a large number of samples ($N=1,000$) generated using the expensive numerical solver with the response of DeepONet trained on the target domain ($N_t=800$) and TL-DeepONet ($N_t=50$). Specifically, in columns 2-5 we show the mean and variance of the probability density of the scalar QoI at two fixed points in space: $\boldsymbol x_1 = (0.323, 0.645)$, $\boldsymbol x_2 = (0.806, 0.258)$ at $t=0.5$. We observe that TL-DeepONet allows for the calculation of the moments of the solution to a satisfactory degree without the need to train the model from scratch with a large number of samples. Furthermore, in columns 6,7 the relative $L_2$ norm error of the mean and variance of the entire evolution of $v$ between MCS and DeepONet, TL-DeepONet is shown. We observe that compared to DeepONet, the error of TL-DeepONet is approximately doubled which results in acceptable error for the mean but high error for the variance of the response, which is expected as capturing higher-order moments is a particularly challenging task. }

\bigbreak
\noindent
\textbf{Conclusions} 

\noindent
Operator regression approaches have been successful in learning nonlinear operators for complex PDEs directly from observations. However, in many real-world applications, collecting the required training data and rebuilding the models is either prohibitively expensive or impossible. In this study, we introduce a novel formulation to transfer the knowledge of a solution operator trained on a system of PDEs for a specific domain to a different domain. Such situations are challenging, considering that the conditional probability distributions of the source and target domains are different. We demonstrate the efficacy of the proposed DeepONet-based transfer learning framework by solving six different benchmark problems (employing well-known differential equations). Our observations can be summarized as follows:
\begin{itemize}
    \item The key ingredients of the proposed approach are the finetuning of TL-DeepONet by freezing lower layers of the network and the formulation of the hybrid loss function in Eq.~\eqref{eq:target-loss}, that aims to reduce the discrepancy of both individual samples and the conditional distribution of the target data. As a rule of thumb, the loss term involving the difference in the conditional probability of the two domains must be penalized more than the regression loss. For easy maneuvering of the appropriate penalizing parameters, we have adopted these parameters to be trainable and self-adaptive during fine-tuning of the target model. \review{From a wide range of experiments performed in this study, we found that the addition of $\cal L_{\text{CEOD}}$ in the hybrid loss function for training of TL-DeepONet improves significantly the predictive accuracy of the model by ensuring the global properties of the conditional distributions of target data are preserved (see Tables \ref{table:TL3_withoutCEOD}, and \ref{table:TL9}).}
    \item \review{Tasks TL1 and TL2 demonstrate the ability of the approach to transfer the knowledge from a square domain to triangular domains with a high accuracy even when small datasets are employed for training (see Table \ref{table:TL1-4}). To test the performance of the approach in challenging situations, we consider domains with discontinuity and notches (tasks TL3 and TL4). We observe a TL-DeepONet accuracy loss of less than $5\%$, demonstrating that one can predict the hydraulic head with very few labelled data $(250)$ even when domains of very different external boundaries are considered. }
    \item The proposed TL framework allows multitask learning even when the source and target domains differ in more than one aspect. As demonstrated in the elasticity model, the two domains are characterized by different internal boundaries and different material properties. A study of the size of the target dataset shows that approximately $200$ samples are sufficient to model the conditional shift from the source to the target domain in TL5 (Table \ref{table:TL5}). \review{However, in TL6, where the internal boundaries and models parameters are changed to a great extent (from smooth to non-smooth boundaries), TL-DeepONet results in relative higher error as the source model is not capable to capture the target features in the lower levels of the network.} 
    \item TL-DeepONet is tested on a dynamical system by employing a 2D time-dependent PDE for a Brusselator reaction-diffusion system. In TL7 we deploy the source model trained on smooth dynamics to be applied for approximating highly non-smooth dynamics. Our results in Table \ref{table:TL5-6} show that even for such challenging dynamics the framework performs well. For TL7, the fine-tuning of the target domain carried out using self-adaptive weights, which was used in the regression loss for the target domain.   
    \item \review{From the reported values of the computational cost, we demonstrate that TL-DeepONet results in significantly faster training than the case of training from scratch the full model with random initialization, even with the same number of training data (of course, training of the source model first with sufficient labeled data is necessary). This result might not seem important when a single TL problem is considered, however it becomes especially important when multiple transfer learning scenarios are considered, whose training from scratch would otherwise lead to a computationally intractable task. Throughout the various applications, we demonstrate that TL-DeepONet can result in very good accuracy even when limited training datasets are considered.}
    \item \review{We found that TL-DeepONet has certain limitations in cases where the complexity (in geometry or model response) is significantly different between the target and the source domain. More specifically, through experiments TL6 and the additional problem TL10 in \ref{sec:add-experiments} (where both internal and external boundaries are modified between source and target domains), we demonstrate that in such cases TL-DeepONet's predictive accuracy deteriorates or in certain cases it requires more training samples, as simply the fine-tuning of the fully-connected layers with very few available samples is not sufficient.}
    \item Overall, we found that transferring previously-acquired knowledge (i.e.,\ domain-invariant features learned from lower levels of the model) and fine-tuning higher-level layers of the network allows for efficient multitask operator learning when solving PDE problems under conditional distribution mismatch. 
\end{itemize}

\section*{Methods}
\bigbreak
\noindent
\textbf{Problem setup:} 
Consider a multi-dimensional function $f(\mathbf{x})$, which represents the mapping between a vector of input random variables, $\mathbf{x}_i \in \mathcal{X}$, and the corresponding output QoIs, $\mathbf{y}_i \in \mathcal{Y}$. Furthermore, consider a source domain for which $N_s$ sufficient labeled data have been generated $\mathcal{D}_s=\{(\mathbf{x}_i^s, \mathbf{y}_i^s)\}_{i=1}^{N_s}$, where $\mathbf{x}_i^{s} \in \mathcal{X}_S$ and $\mathbf{y}_i^{s} \in \mathcal{Y}_S$. In addition, there exists a target domain with very few $N_t$ available labeled data, and a set of $N_u$ additional unlabeled data, i.e., $\mathcal{D}_t=\{(\mathbf{x}_i^{tL}, \mathbf{y}_i^{tL})\}_{i=1}^{N_t} \cup \{\mathbf{x}_j^{tU}\}_{j=1}^{N_u}$, where $\mathbf{x}_i^{tL}, \mathbf{x}_j^{tU} \in \mathcal{X}_S$ and $\mathbf{y}_i^{tL} \in \mathcal{Y}_T$. Thus, we focus on transfer learning under conditional shift, where the marginal distributions are identical ($\mathcal{X}_S$) and the conditional distributions differ ($\mathcal{Y}_S \ne \mathcal{Y}_T$).

An important task in TL is to accurately compute the discrepancy between conditional distributions. While domain adaptation problems focus on minimizing the discrepancy between source and target distributions, here we aim to address the distribution shift between the labeled target data $\{\mathbf{y}_i^{tL}\}_{i=1}^{N_t}$ and the surrogate prediction on the unlabeled target data $\{f_T(\mathbf{x}_i^{tU})\}_{i=1}^{N_u}$. Several approaches have been proposed to measure the discrepancy or \textit{divergence} between the marginal and conditional distributions. Early methods relied on the estimation of the underlying distributions via density estimation methods, such as kernel density estimation (KDE) \cite{lee2006estimation}. However, when high-dimensional real-world data are considered, density estimation can be very challenging, and thus modern approaches avoid this intermediate step altogether. Yu et al.\ \cite{yu2020measuring}, proposed a new statistic, which operates on the cone of a symmetric positive semidefinite (SPS) matrix using the Bregman matrix divergence and used the cross entropy function to explicitly incorporate higher order information in the data. An alternative approach is to compute the distance metric through the embedding of probability measures in a reproducing kernel Hilbert space (RKHS). A RKHS is a Hilbert space where all evaluation functionals in it are bounded \cite{muandet2017kernel}. In such methods, the difference between the mean embedding in RKHS is computed like in  Maximum Mean Discrepancy (MMD) proposed by Gretton et al.\ \cite{gretton2012kernel}, which operates on marginal distributions. In this work, we employ a measure developed for conditional distributions, and thus below we briefly present some necessary preliminaries of the conditional embedding operator (CEO) theory.
\bigbreak
\noindent
\textbf{Conditional embedding operator theory} :
Consider two random variables $X$ and $Y$, with $\Omega_X, \Omega_Y$ and $\mathcal{H}, \mathcal{F}$ being the original and RKHS spaces, respectively. The conditional mean embedding of $P(Y\vert X)$ can be defined as \cite{song2013kernel}:
\begin{equation}
\label{eq:conditional-mean-emb_a}
\mu_{Y\vert\mathbf{x}} := \mathbb{E}_{Y\vert\mathbf{x}}[\psi(Y)\vert X=\mathbf{x}] = \mathcal{C}_{Y\vert X} \phi(\mathbf{x}),
\end{equation}
which must satisfy the reproducibility requirement such that:
\begin{equation}
\label{eq:conditional-mean-emb_b}
\mathbb{E}_{Y\vert \mathbf{x}}[\psi(Y)\vert X=\mathbf{x}] = \langle h, \mu_{Y\vert \mathbf{x}} \rangle_{\mathcal{F}} 
\ \ \ \forall \;\; h \in \mathcal{F},
\end{equation}
where $\phi(\mathbf{x}): \Omega_X \rightarrow \mathcal{H}$, $\psi(Y): \Omega_Y \rightarrow \mathcal{F}$ and $\mathcal{C}_{Y\vert X}$ is an operator $\mathcal{H} \rightarrow \mathcal{F}$.

Unlike the case of a marginal distribution, the embedding of a conditional distribution does not result in a single element in the RKHS but rather a family of points, each indexed by a fixed value $\mathbf{x}$ of the conditioning variable $X$. Thus, to obtain a single RKHS element $\mu_{Y\vert \mathbf{x}} \in \mathcal{F}$, we need to fix $X$ to a particular value $\mathbf{x}$. Therefore, the operator $\mathcal{C}_{Y\vert X}$ takes as input a given $\mathbf{x}$ and outputs an embedding. Based on Song et al.\ \cite{song2009hilbert}, the operator $\mathcal{C}_{Y\vert X}$ is defined as
\begin{equation}
\label{eq:operator}
\mathcal{C}_{Y\vert X} = \mathcal{C}_{YX} \inv{\mathcal{C}_{XX}} =  
\mathbb{E}_{YX}[\psi(Y) \otimes \phi(X)] \inv{\mathbb{E}_{XX}}[\psi(X) \otimes \phi(X)],
\end{equation}
where  $\mathcal{C}_{YX}$ and $\mathcal{C}_{XX}$ are the cross-covariance and self-covariance operators, respectively. Given an available dataset $\mathcal{D}=\{(\mathbf{x}_i, \mathbf{y}_i)\}_{i=1}^{N}$, the empirical estimator of the operator $\mathcal{C}_{Y\vert X}$ can be computed as
\begin{equation}
\label{eq:emp-operator}
\hat{\mathcal{C}}_{Y\vert X} = \hat{\mathcal{C}}_{YX} \inv{\hat{\mathcal{C}}_{XX}} = \Phi\inv{(\mathbf{K} + \lambda N \mathbf{I})} \Upsilon^\top.
\end{equation}
\review{where $N$ is the number of samples, $\Phi := (\psi(\mathbf{y}_1), \dots \psi(\mathbf{y}_N))$, $\Upsilon := (\phi(\mathbf{x}_1), \dots \phi(\mathbf{x}_N))$ and $\mathbf{K} = \Upsilon^\top \Upsilon$ is the Gram matrix for the samples generated from $X$. To avoid overfitting, an additional regularization parameter $\lambda$ is added.}

Given two datasets $\mathcal{D}_p=\{(\mathbf{x}_i, \mathbf{y}_i)\}_{i=1}^{N_1}$ and $\mathcal{D}_q=\{(\mathbf{x}_i, \mathbf{y}_i)\}_{i=1}^{N_2}$, MMD measures the discrepancy between the mean embeddings  using a Hilbert-Schmidt norm as $D_{\text{MMD}}(\mathcal{D}_p, \mathcal{D}_q) = \| \hat{\mu}_{X_p} - \hat{\mu}_{X_q} \|^{2}_{HS}$. Inspired by the MMD, Liu et al.\ \cite{liu2021deep}, recently proposed the Conditional Embedding Operator Discrepancy (CEOD) to measure the divergence between conditional distributions. The COED is based on empirical conditional embedding operators and is constructed as
\begin{equation}
\label{eq:ceod}
\begin{aligned}
D_{\text{CEOD}}(\mathcal{D}_p, \mathcal{D}_q) &= \left\|\hat{\mathcal{C}}_{Y_p\vert X_p} - \hat{\mathcal{C}}_{Y_q\vert X_q}\right\|_{HS}^2 \\
&= \|\Phi(Y_p) \inv{(\mathbf{K}_{X_p X_p} + \lambda N_1 \mathbf{I})} \Upsilon^\top (X_p)\\
&- \Phi(Y_q) \inv{(\mathbf{K}_{X_q X_q} +\lambda N_2 \mathbf{I})} \Upsilon^\top (X_q)\|_{HS}^2 \\
&= \text{Tr}\Big\{ \inv{(\mathbf{K}_{X_p X_p} + \lambda N_1 \mathbf{I})} \mathbf{K}_{Y_p Y_p} \inv{(\mathbf{K}_{X_p X_p} + \lambda N_1 \mathbf{I})} \mathbf{K}_{X_p X_p} \Big\} \\
&+ \text{Tr}\Big\{ \inv{(\mathbf{K}_{X_q X_q} + \lambda N_2 \mathbf{I})} \mathbf{K}_{Y_q Y_q} \inv{(\mathbf{K}_{X_q X_q} + \lambda N_2 \mathbf{I})} \mathbf{K}_{X_q X_q} \Big\} \\
&- 2\text{Tr}\Big\{ \inv{(\mathbf{K}_{X_p X_p} + \lambda N_1 \mathbf{I})} \mathbf{K}_{Y_p Y_q} \inv{(\mathbf{K}_{X_q X_q} + \lambda N_2 \mathbf{I})} \mathbf{K}_{X_q X_p} \Big\},
\end{aligned}
\end{equation}
where $\mathbf{K}_{XX'}(i,j) = k(\mathbf{x}_i, \mathbf{x}'_j)$ is the Gram matrix calculated with a Gaussian kernel $k$. 

We note that methods that rely on kernel embeddings may be expensive. For this reason, a low-rank approximation of the Gram matrix, such as incomplete Cholesky factorization, can be used, which allows the reduction of computational cost while maintaining sufficient precision \cite{song2013kernel}. In this work, the CEOD is used as part of the hybrid loss used to train the target surrogate model after transferring trained variables from the source surrogate as elucidated in the sequel.


\vspace{0.1in} 
\noindent{\bf Steps of the proposed method} \\
\noindent
\textbf{DeepONet training on source domain}:
A DeepONet surrogate is trained on a source domain $\mathcal{D}_s=\{(\mathbf{x}_i^s, \mathbf{y}_i^s)\}_{i=1}^{N_s}$, where $\mathbf{x}_i^{s} \in \mathcal{X}_S$ and $\mathbf{y}_i^{s} \in \mathcal{Y}_S$. Any neural network architecture can be chosen for the branch net and the trunk net of the DeepONet to encode the input function and spatio-temporal coordinates, respectively. The source model is trained with a standard regression loss $\mathcal L_r(\theta^S)$, such as the relative $L_2$ error
\begin{equation}
\label{eq:source-loss}
    \mathcal L(\theta^S) = \mathcal L_r(\theta^S) = \frac{\|f_S(\mathbf{x}^{s})- {\mathbf{y}^s\|}_{2}}{{\|\mathbf{y}^s\|}_{2}},
\end{equation}
where ${\|\cdot\|}_2$ denotes the standard Euclidean norm and $f_S(\mathbf{x}^{s}), \mathbf{y}^s$ are the prediction and reference responses,  respectively. Alternative error measures such as the mean-square error (MSE) can also be employed. The loss minimization is performed using the \review{gradient descent algorithm, particularly the Adam optimizer}. After training, the solution operator $\mathcal G_{\theta^S}$ is learned and the trained parameters of the network $\theta^{S*}$ are saved. In Figure \ref{fig:DeepONet}, we present a schematic of the overall approach, where the source training is depicted inside the blue box.
\bigbreak
\noindent
\textbf{Parameters transfer}:
The trained parameters of DeepONet are transferred to the target model, referred to as TL-DeepONet, for better initialization. As discussed previously, domain-invariant features transferred to the target model provide better initialization of the network, which alleviates the computational cost of training from scratch with random initialization. Furthermore, this approach addresses the issue of overfitting when very few target samples are available \cite{saxe2019information}. Additionally, the architecture of TL-DeepONet (both the branch and the trunk network) remains identical. Certain layers of it are frozen while others are fine-tuned as described next. In Figure \ref{fig:DeepONet}, we present the training of TL-DeepONet after the transferring of the trained source parameters $\theta^{S*}$ (red box).
\bigbreak
\noindent
\textbf{TL-DeepONet fine-tuning on target domain}:
As discussed previously, training the surrogate with scarce target data can lead to overfitting. Hence, certain task-specific layers of the network are fine-tuned while others remain frozen with constant parameters. Commonly, in computer vision, it is accepted that the convolutional layers are general, while fully-connected layers are task-specific \cite{yosinski2014transferable, liu2021deep}. Adapting this concept in \review{deep neural networks (DNN)}, we propose the fine-tuning of the fully-connected network of the branch CNN $\{ f^l_{b_1}, \dots f^l_{b_m}\}$, where $m$ is the total number of layers, and the last layer of the trunk net $f^l_{t}$, to allow for sufficient expressivity during the training of the target DeepONet while maintaining the low training cost. In Figure \ref{fig:DeepONet}, the task-specific layers are denoted with red arrows. 
\bigbreak
\noindent
\textbf{Hybrid loss function}:
Finally, we train the TL-DeepONet using a hybrid loss $\mathcal L(\theta^T)$, which considers not only the accurate match between individual target samples but also the agreement between the conditional distributions of the target data. The regression loss $\mathcal L_r(\theta^T)$ is simply computed as in Eq.~\eqref{eq:source-loss} for the target domain data $\mathcal{D}_t=\{(\mathbf{x}_i^{tL}, \mathbf{y}_i^{tL})\}_{i=1}^{N_t}$. To compute the CEOD loss $\mathcal L_{\text{CEOD}}(\theta^T)$ \cite{liu2021deep}, we consider as input data the output of the first fully-connected layer of the branch net $f^l_{b_1}$, which we denote as $\mathbf{x}_{b_1}$. Then, we measure the conditional distribution discrepancy (Eq.~\eqref{eq:ceod}) between the labeled data $\mathcal{D}^L_t=\{(\mathbf{x}_{b_1 i}^{tL}, \mathbf{y}_i^{tL})\}_{i=1}^{N_t}$ and unlabeled data $\mathcal{D}^U_t=\{(\mathbf{x}_{b_1 i}^{tU}, f_T(\mathbf{x}_i^{tU})\}_{i=1}^{N_u}$. Considering both components, the hybrid loss reads
\begin{equation}
\label{eq:target-loss}
\begin{aligned}
    \mathcal L(\theta^T) &= \lambda_1 \mathcal L_r(\theta^T) + \lambda_2 \mathcal L_{\text{CEOD}}(\theta^T) \\ &=  \lambda_1 \frac{\|f_T(\mathbf{x}^{tL})- {\mathbf{y}^{tL}\|}_{2}}{{\|\mathbf{y}^{tL}\|}_{2}} + \lambda_2 \left\|\hat{\mathcal{C}}_{Y_{tL}\vert X_{tL}} - \hat{\mathcal{C}}_{Y_{tU}\vert X_{tU}}\right\|_{HS}^2
\end{aligned}
\end{equation}
where $\lambda_1 = 1$ and 
$\lambda_2>>\lambda_1$ are trainable coefficients, which determine the importance of the two loss components during the optimization process. The trainable coefficients are updated during backpropagation. During fine-tuning of the target model, we simultaneously minimize the loss function, $\mathcal L(\theta^T)$, with respect to $\theta^T$, and maximize the loss function with respect to $\lambda_2$. This approach progressively penalizes the target network for the discrepancy in the conditional probabilities of the two tasks. An important aspect of this approach is the initialization of $\lambda_2$ at the beginning of optimization, which is problem specific. For all the problems presented in this work, we have initialized $\lambda_2 = 10$ at the beginning of the training. By training TL-DeepONet based on the hybrid loss in Eq.~\eqref{eq:target-loss}, we obtain the optimized parameters $\theta^{T*}$. 

\backmatter

\subsection*{Data and Code Availability}
The code and dataset generation scripts used in this study is available on the GitHub repository \url{https://github.com/katiana22/TL-DeepONet}.

\subsection*{Acknowledgments}

For K.K. and M.D.S., this material is based upon work supported by the U.S. Department of Energy, Office of Science, Office of Advanced Scientific Computing Research under Award Number DE-SC0020428. S.G. and G.E.K. would like to acknowledge support by the DOE project PhILMs (Award Number DE-SC0019453) and the OSD/AFOSR MURI grant FA9550-20-1-0358.

\subsection*{Author Contributions Statement}
\noindent
S.G.: conceptualization, data curation, formal analysis, investigation, methodology, software, validation, visualization, writing -- original draft, writing -- review and editing.\\
K.K.: conceptualization, data curation, formal analysis, investigation, methodology, software, validation, visualization, writing -- original draft, writing -- review and editing.\\
M.D.S.: conceptualization, funding acquisition, investigation, project administration, resources, supervision, writing -- original draft, writing -- review \& editing.\\
G.E.K.: conceptualization, funding acquisition, project administration, resources, supervision, writing---original draft, writing -- review \& editing.

\subsection*{Competing Interests Statement}
The authors declare no competing interest.

\newpage
\makeatletter
\renewcommand \thesection{S\@arabic\c@section}
\renewcommand\thetable{S\@arabic\c@table}
\renewcommand \thefigure{S\@arabic\c@figure}
\makeatother
\section*{Supplementary information}
\setcounter{figure}{0}
\setcounter{table}{0}
\setcounter{section}{0}
\setcounter{page}{1}

\section{Theoretical details}
\label{sec:learning-scenario} 

\bmhead{Detailed description of the transfer learning scenario} 
Consider a nonlinear and high-fidelity partial differential equation (PDE) model describing a physical process and a corresponding (usually expensive) numerical simulator (e.g., a finite-difference or finite-element solver). In a standard UQ setting, our aim is to approximate the mapping $f(\mathbf{x})$, between a vector of input random variables, $\mathbf{x}_i \in \mathcal{X}$, and the corresponding output quantities of interest (QoIs), $f(\mathbf{x}_i) = \mathbf{y}_i \in \mathcal{Y}$. We denote the input realizations (or instances) as $\mathbf{X} = \{\mathbf{x}_{1},..,\mathbf{x}_{N}\}$ and the associated QoIs as $\mathbf{Y} = \{\mathbf{y}_{1},..,\mathbf{y}_{N}\}$. 
Thus, model $f$ performs the mapping $f:{\mathbf{x}_i \in \mathbb{R}^{D_{\text{in}}}} \rightarrow \mathbf{y}_i \in {\mathbb{R}^{D_{\text{out}}}}$ where the dimensionality of both the inputs and the outputs, $D_{\text{in}}, D_{\text{out}}$ is usually high, e.g., $\mathcal{O}(10^{2-4})$. Here, the high-dimensional inputs may represent random fields and/or processes, such as spatially or temporally varying coefficients, and the corresponding QoIs represent physical quantities, which can also vary in both space and time. In this regression setting, we aim to approximate the mapping, $f$, based on a training dataset of $N$ input-output pairs, ($\mathbf{X} = \{\mathbf{x}_{1},..,\mathbf{x}_{N}\}$, $\mathbf{Y} = \{\mathbf{y}_{1},..,\mathbf{y}_{N}\}$), and achieve the lowest possible predictive error.

Suppose that a surrogate $f_S$ is learned on a \textbf{source domain}, i.e., under specified conditions a dataset with $N_s$ sufficient labeled data is generated $\mathcal{D}_s=\{(\mathbf{x}_i^s, \mathbf{y}_i^s)\}_{i=1}^{N_s}$. The generated dataset is composed of two parts: the random input realizations $ \mathbf{x}_i^s \in \mathcal{X}_S$ and the corresponding labels  $ \mathbf{y}_i^s \in \mathcal{Y}_S$. Now consider a second \textbf{target domain} corresponding to different problem conditions. These conditions might represent the simulation box geometry, boundary conditions, model parameters, etc. Let us assume that only few available labeled target data exist in $\mathcal{D}_t=\{(\mathbf{x}_i^t, \mathbf{y}_i^t)\}_{i=1}^{N_t}$, where $\mathbf{x}_i^t \in \mathcal{X}_T$, $ \mathbf{y}_i^t \in \mathcal{Y}_T$ and $N_t \ll N_s$. Furthermore, we assume that there exists a conditional shift, under which the marginal distribution of the inputs remains the same $P(\mathbf{x}_s) = P(\mathbf{x}_t)$, while the conditional distributions of the QoIs are different $P(\mathbf{y}_s\vert \mathbf{x}_s) \ne P(\mathbf{y}_t\vert \mathbf{x}_t)$. In cases where multiple target tasks exist, collecting sufficient labeled data can be computationally prohibitive. Furthermore, training a surrogate for target tasks with scarce data can lead to overfitting. Therefore, in this work we aim to construct fast and efficient surrogates $f_T$ for target tasks by leveraging and transferring information between different but related domains, $\mathcal{D}_s$ and $\mathcal{D}_t$, without the need to retrain models from scratch with random initialization. As a surrogate model we will employ the deep neural operator or DeepONet \cite{lu2021learning}, a neural network-based method, which enables task-specific learning of PDE operators. More information on the DeepONet is given in the next Section. 

\bmhead{Deep operator network (DeepONet)}

Deep neural networks (DNNs) have been demonstrated to be an effective and versatile tool for problems where the solutions are ambiguous or where there is insufficient knowledge about the relationships between the inputs and the outputs. The ability of DNNs to approximate arbitrary continuous functions in compact domains is a key advantage. On the other hand, training DNNs is done for differential equations with fixed inputs, such as initial conditions, boundary conditions, forcing, and coefficients. If one of the inputs is changed, the training process must be restarted. It is difficult to obtain real-time outputs for multiphysics systems that require multiple sets.  

Operator learning, which is inspired by the universal approximation theorem of operators, can be used to overcome the limitation of functional regression. DeepONet's seminal work on learning diverse continuous nonlinear operators motivated our work. DeepONet is particularly influenced by theory that guarantees a small approximation error (that is, the error between the target operator and the class of neural networks of a given finite-size architecture). Before delving into the solution operators of parametric PDEs, it is critical to understand the distinction between function regression and operator regression. The solution operator in the function regression approach is parametrized as a neural network between finite-dimensional Euclidean spaces: $C:\mathbb{R}^{d_1}\to\mathbb{R}^{d_1}$, where $d_1$ is the number of discretization points. In operator regression, however, a function is mapped to another function using an operator. In other words, it is the transformation of one infinite-dimensional space into another infinite-dimensional space. By learning the non-linear operator from the data, the operators would be trained to approximate the solution of the input functions using operator regression.

The DeepONet architecture consists of two DNNs: the branch net encodes the input function, $\mathbf X$, at fixed sensor points, $\{x_1, x_2, \dots, x_m\}$, while the trunk net encodes the information related to the spatio-temporal coordinates, $\zeta = \{x_i, y_i, t_i\}$, at which the solution operator is evaluated to compute the loss function. When a physical system is described by PDEs, it involves multiple functions, for example, the PDE solution, $\boldsymbol u(x,t)$, the forcing term, $\boldsymbol f(x, t)$, the initial condition, $\boldsymbol u_0(x)$, and the boundary conditions, $\boldsymbol u_b(x, t)$, where $x$ and $t$ are the spatial and temporal coordinates, respectively. We are usually interested in one of these functions, which is the output of the solution operator, and try to predict it based on the varied forms of the other functions, which is the input to the branch net. The trunk net takes as input the spatial and the temporal coordinates, e.g. $\zeta = \{x_i, y_i, t_i\}$, at which the solution operator is evaluated to compute the loss function.

Consider a computational model, $\mathcal{M}(\mathbf{x})$, where $\mathbf{x}_i=\{\mathbf{x}_{i}(x_1), \mathbf{x}_{i}(x_2), \ldots, \mathbf {x}_{i}(x_m)\}$ (pointwise evaluation of the input function to the branch net), which simulates a physical process and represents a mapping between a vector of input random variables, $\mathbf{x}(\zeta) \in \mathbb{R}^{D_{\text{in}}}$, and the corresponding output quantities of interest (QoIs), $\mathbf{y}(\zeta) \in \mathbb{R}^{D_{\text{out}}}$ where $\zeta$ represent spatio-temporal coordinates. The goal of the DeepONet is to learn the solution operator, $\mathcal G(\mathbf{x})$ that approximates $\mathcal{M}(\mathbf{x})$, and can be evaluated at continuous spatio-temporal coordinates, $\zeta$ (input to the trunk net). The output of the DeepONet for a specified input vector, $\mathbf{x}_i$, is a scalar-valued function of $\zeta$ expressed as $\mathcal G_{\boldsymbol\theta}(\mathbf{x}_i)(\zeta)$, where $\boldsymbol{\theta} = \left(\mathbf W, \mathbf b \right)$ includes the trainable parameters (weights, $\mathbf W$, and biases, $\mathbf b$) of the networks.

 The solution operator for an input realization, $\mathbf x_1$, can be expressed as: 
\begin{equation}\label{eq:output_deeponets}
    \begin{split}
      \mathcal G_{\boldsymbol \theta}(\mathbf{x}_1)(\zeta) &= \sum_{i = 1}^p b_i \cdot tr_i = \sum_{i = 1}^{p}b_i(\mathbf{x}_{1}(x_1), \mathbf{x}_{1}(x_2), \ldots, \mathbf {x}_{1}(x_m))\cdot tr_i(\zeta),   
\end{split}
\end{equation}
where ${b_1, b_2, \ldots, b_p}$ are outputs of the branch net and ${tr_1, tr_2, \ldots, tr_p}$ are outputs of the trunk net. Conventionally, the trainable parameters of the DeepONet, represented by $\boldsymbol{\theta}$ in Eq.~\eqref{eq:output_deeponets}, are obtained by minimizing a loss function, which is expressed as:
\begin{equation}
    \mathcal L(\boldsymbol{\theta}) = \mathcal L_r(\boldsymbol{\theta}) + \mathcal L_i(\boldsymbol{\theta}),
\end{equation}
where $\mathcal L_r(\boldsymbol{\theta})$ and $\mathcal L_i(\boldsymbol{\theta})$ denote the residual loss and the initial condition loss, respectively. 

The DeepONet model provides a flexible paradigm that does not limit branch and trunk networks to any particular architecture. For an equispaced discretization of the input function, a convolutional neural network (CNN) could be used for the branch net architecture, while for a sparse representation of the input function, one could also use a feedforward neural network (FNN). A standard practice is to use a FNN for the trunk network to take advantage of the low dimensions of the evaluation points, $\zeta$.

Although the original DeepONet architecture proposed in \cite{lu2021learning} has shown remarkable success, several extensions have been proposed in \cite{lu2021comprehensive} and \cite{kontolati2022influence} to modify its implementation and produce efficient and robust architectures. For instance, in the POD-DeepONet the basis functions for the trunk net are computed by performing proper orthogonal decomposition (POD) on the training data and using these basis in place of the trunk net. Additionally, the self-adaptive weights in the loss function of the DeepONet helps to automatically maneuver for the optimal penalizing parameter for different terms in the loss function to approximate systems with steep gradients.

\section{Network architecture details}
\label{sec:network-details} 

We list the architectures of DeepONet for each example in Table~\ref{table:architectures}. The DeepONet architecture used for the target domain is identical to the DeepONet architecture used for training the source domain. 

\begin{table}[h!]
\footnotesize
\caption{DeepONet architectures for the transfer learning problem (TL1-TL8).}
\centering
\begin{tabular}{c c c c}
\toprule
 Model & Branch net & Trunk net & \begin{tabular}{@{}c@{}}Activation \\ function\end{tabular}  \\
 \toprule
TL1, TL2 \& TL3   & CNN $+\;[512, 256, 150]$ & $[2, 128, 128, 128, 150]$ & Leaky ReLU  \\
TL4  &  CNN $+\;[256, 512, 150]$ & $[2, 128, 128, 128, 150]$ & Leaky ReLU \\  
\multirow{2}{*}{TL5} & Linear: $[101,100]$ & \multirow{2}{*}{$[2, 128, 128, 128, 150]$} & \multirow{2}{*}{Leaky ReLU}\\
& CNN $+\;[512, 512, 150]$ & \\
\multirow{2}{*}{TL6} & Linear: $[101,100]$ & \multirow{2}{*}{$[2, 128, 128, 128, 200]$} & \multirow{2}{*}{Leaky ReLU}\\
& CNN $+\;[256, 512, 200]$ & \\
TL7 \& TL8  &  CNN $+\;[512, 512, 150]$ & $[2, 128, 128, 128, 150]$ & Leaky ReLU \\
\bottomrule
\end{tabular}
\label{table:architectures}
\end{table}

\section{Data generation}
\label{sec:Data generation} 
\label{sec:data_generation}

\begin{figure}[h!]
\centering
\subfigure[]{
\includegraphics[width=0.23\textwidth, trim= 6cm 7.5cm 6cm 7.5cm, clip]{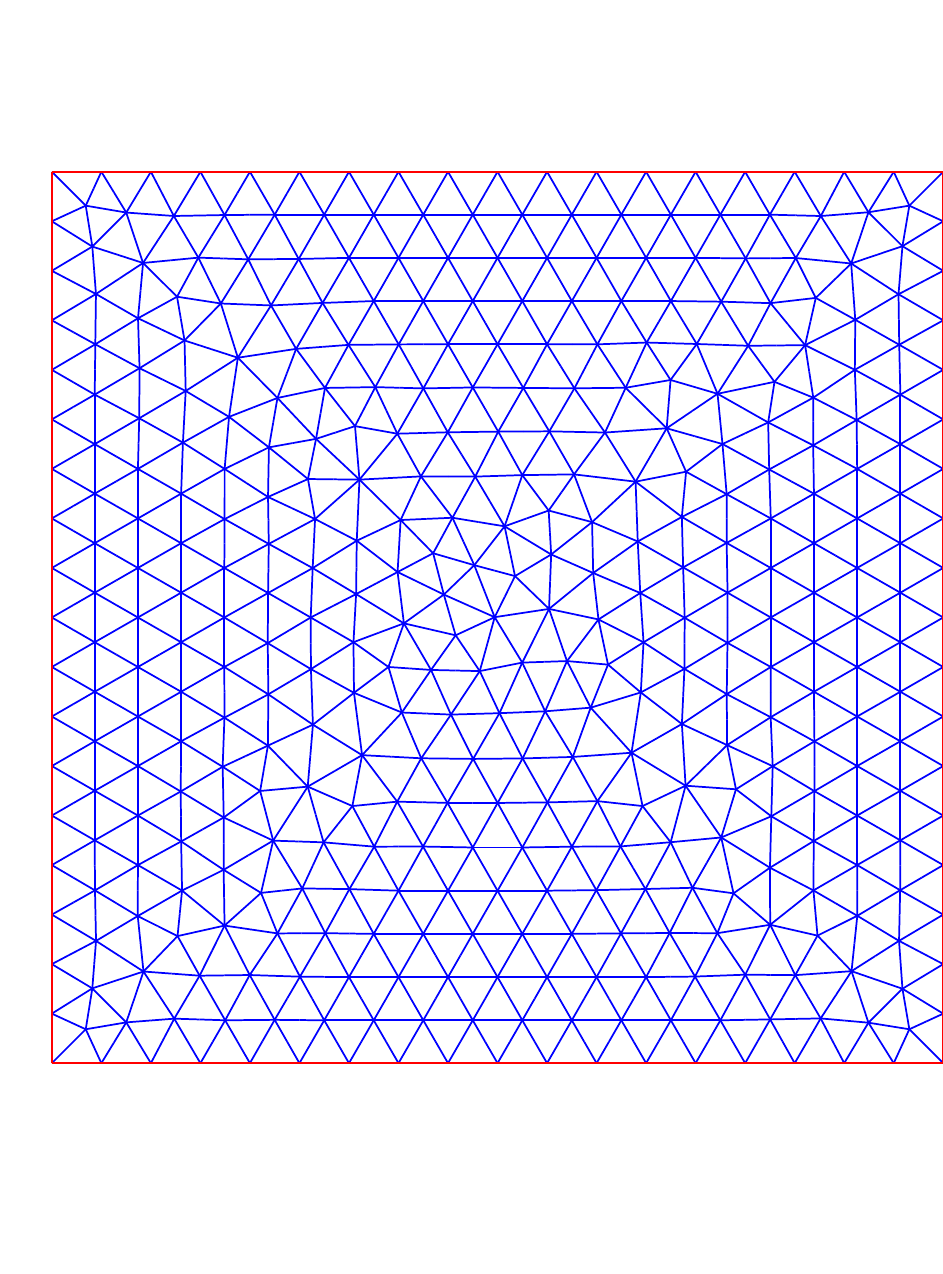}}
\subfigure[]{
\includegraphics[width=0.23\textwidth, trim= 5.8cm 7.5cm 5cm 9.3cm, clip]{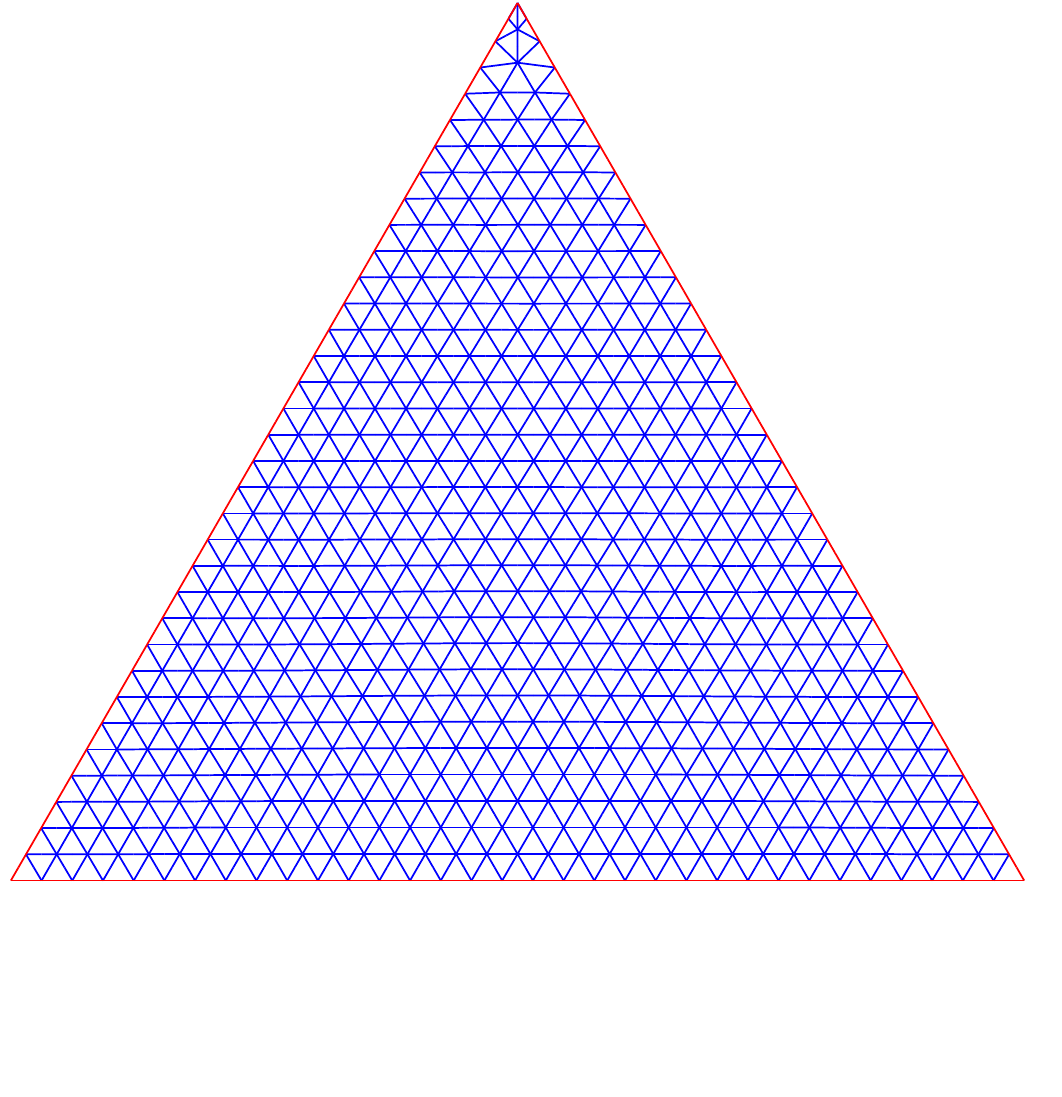}}
\subfigure[]{
\includegraphics[width=0.23\textwidth, trim= 6cm 7.5cm 6cm 7.5cm, clip]{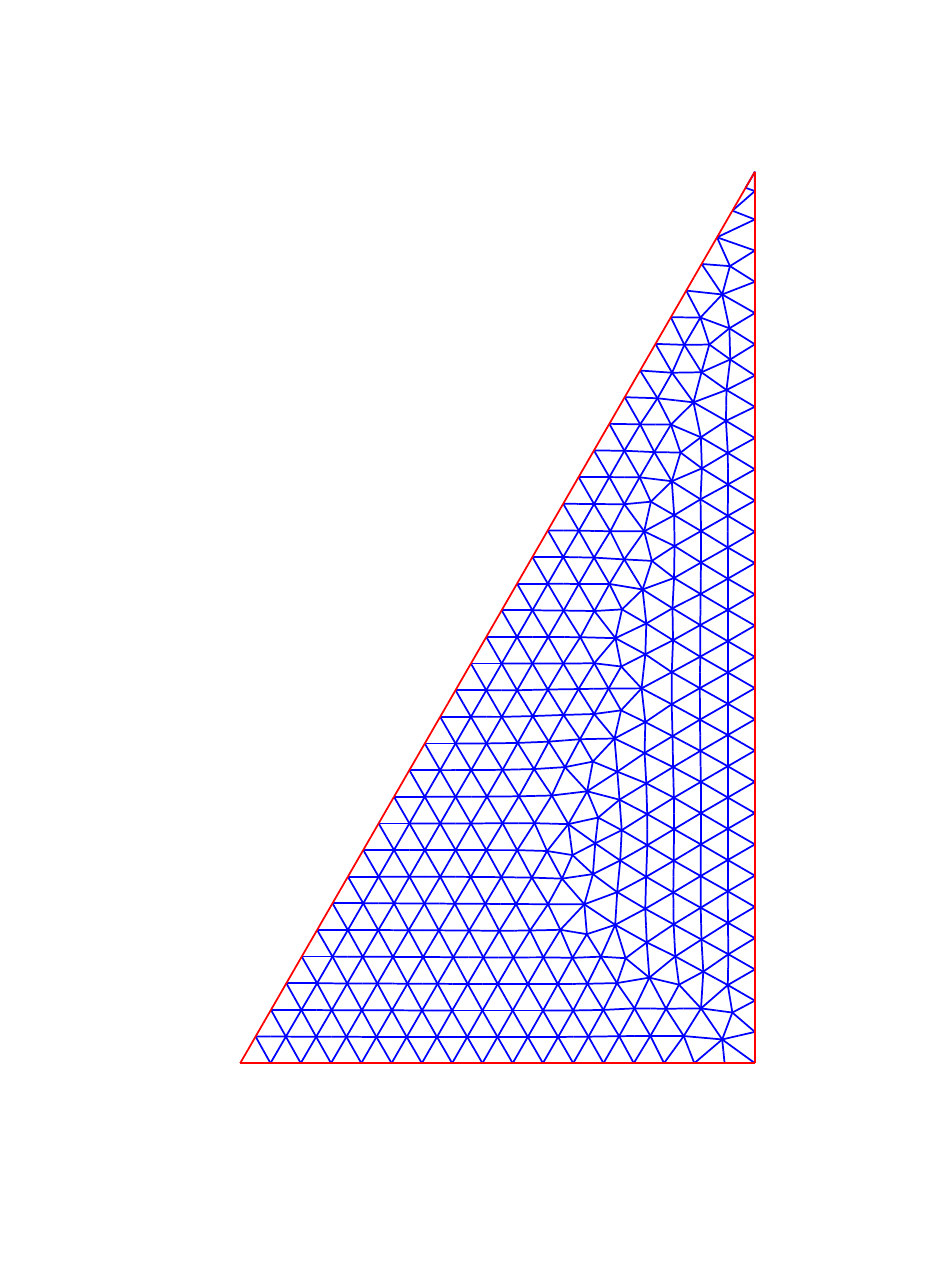}}
\subfigure[]{
\includegraphics[width=0.23\textwidth, trim= 5.5cm 7.5cm 5.2cm 7.5cm, clip]{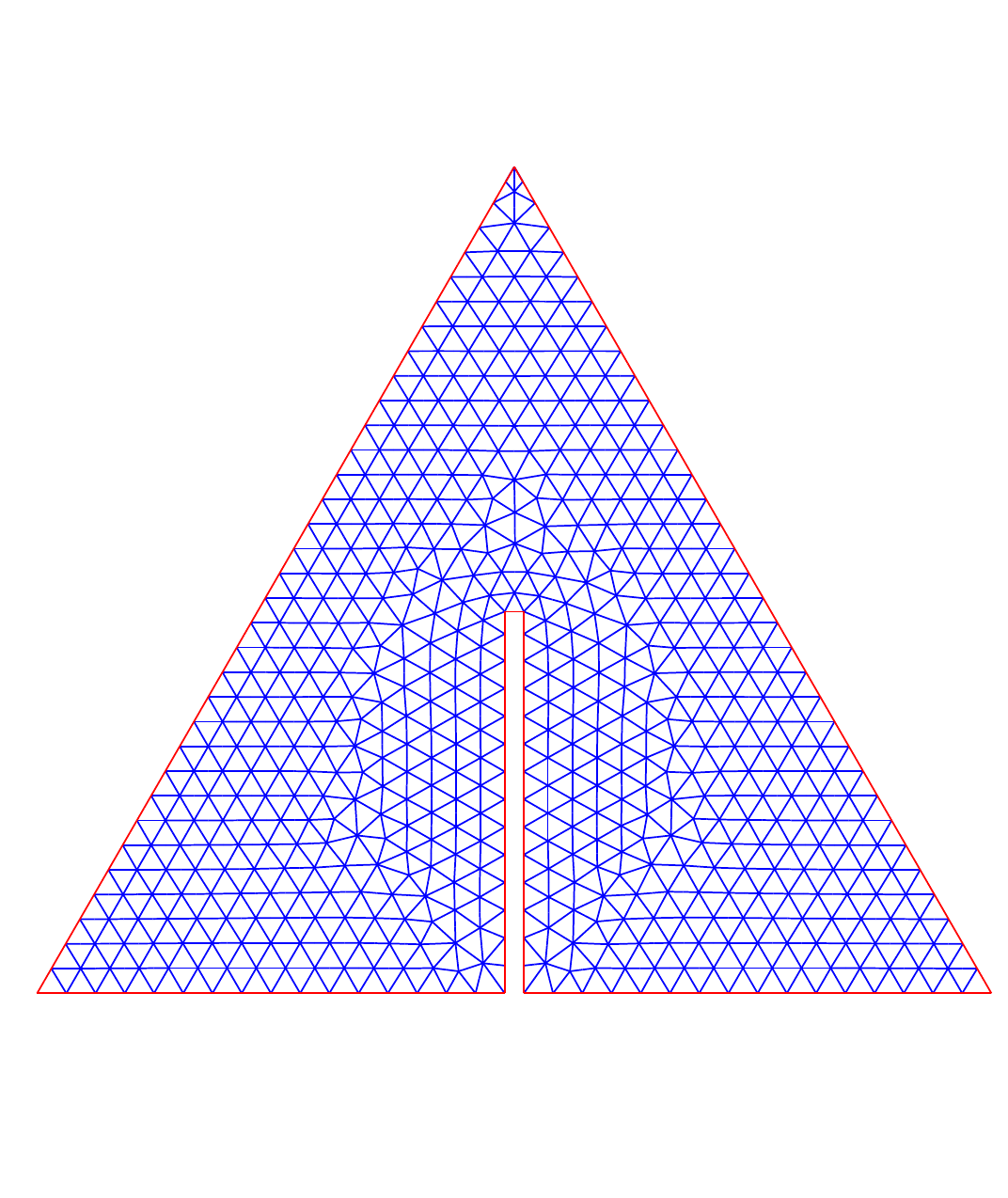}}
\subfigure[]{
\includegraphics[width=0.23\textwidth, trim= 6cm 7.5cm 6cm 7.5cm, clip]{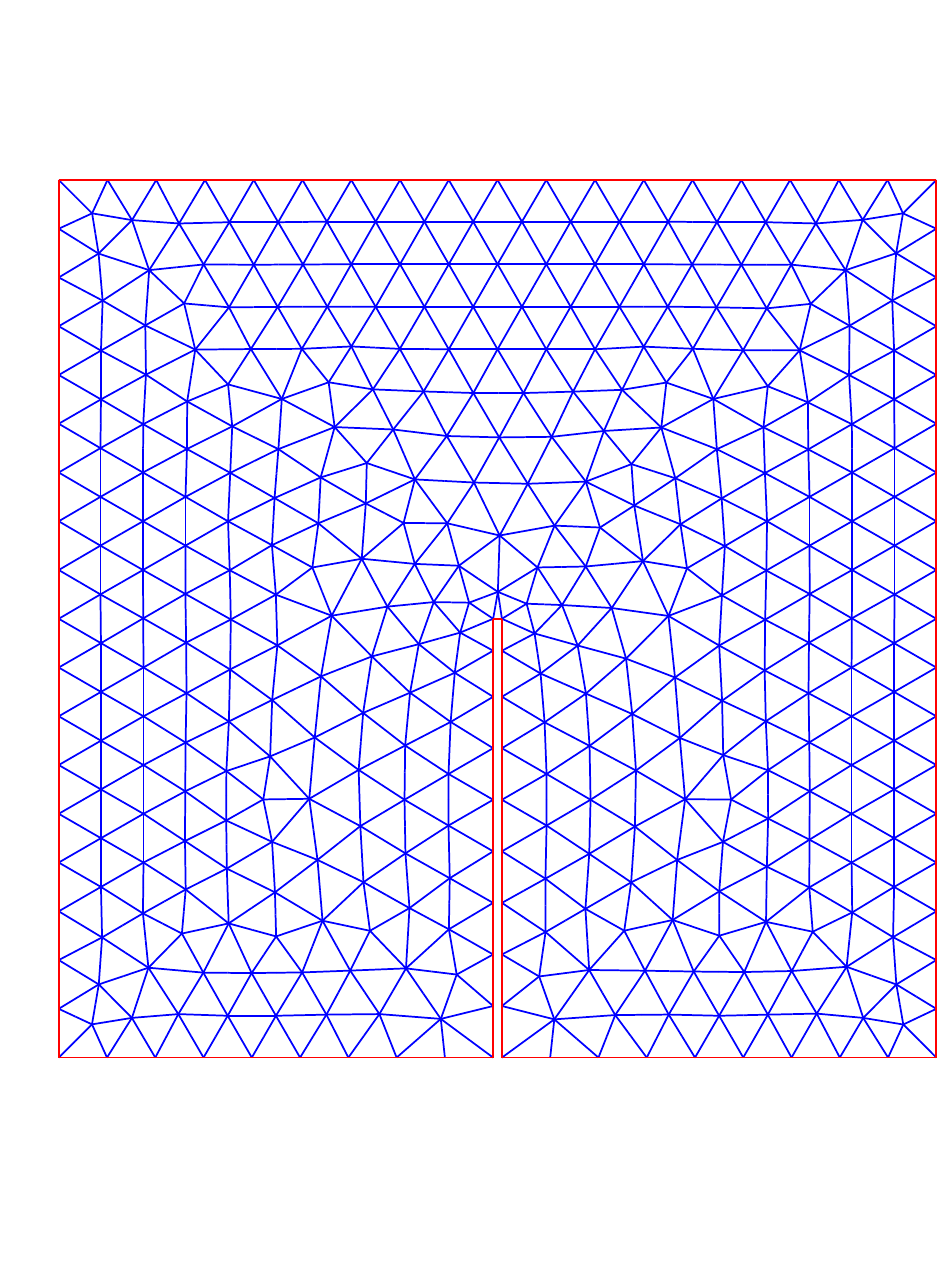}}
\subfigure[]{
\includegraphics[width=0.23\textwidth, trim= 6cm 7.5cm 6cm 7.5cm, clip]{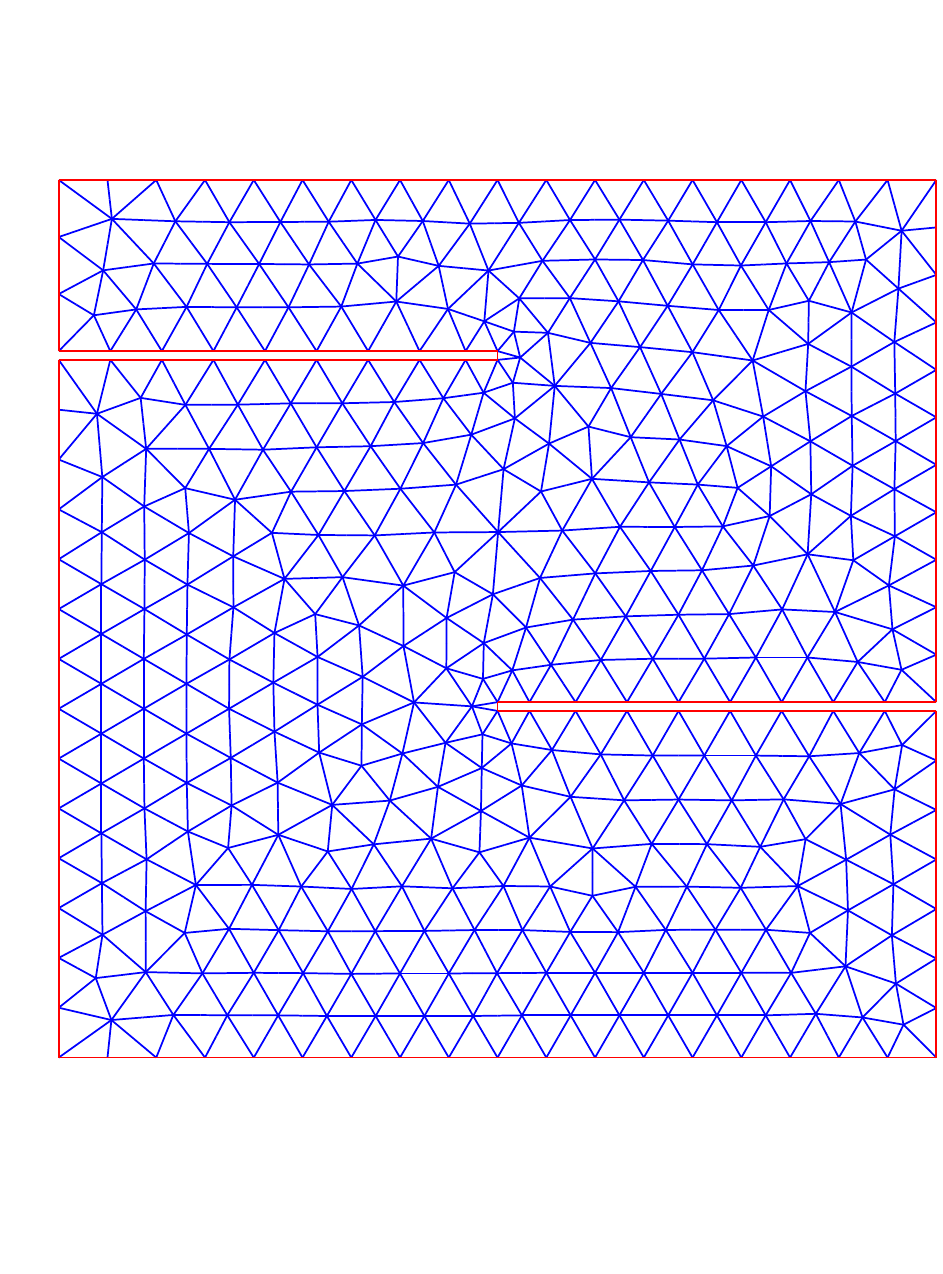}}
\subfigure[]{
\includegraphics[width=0.23\textwidth, trim= 6cm 7.5cm 6cm 7.5cm, clip]{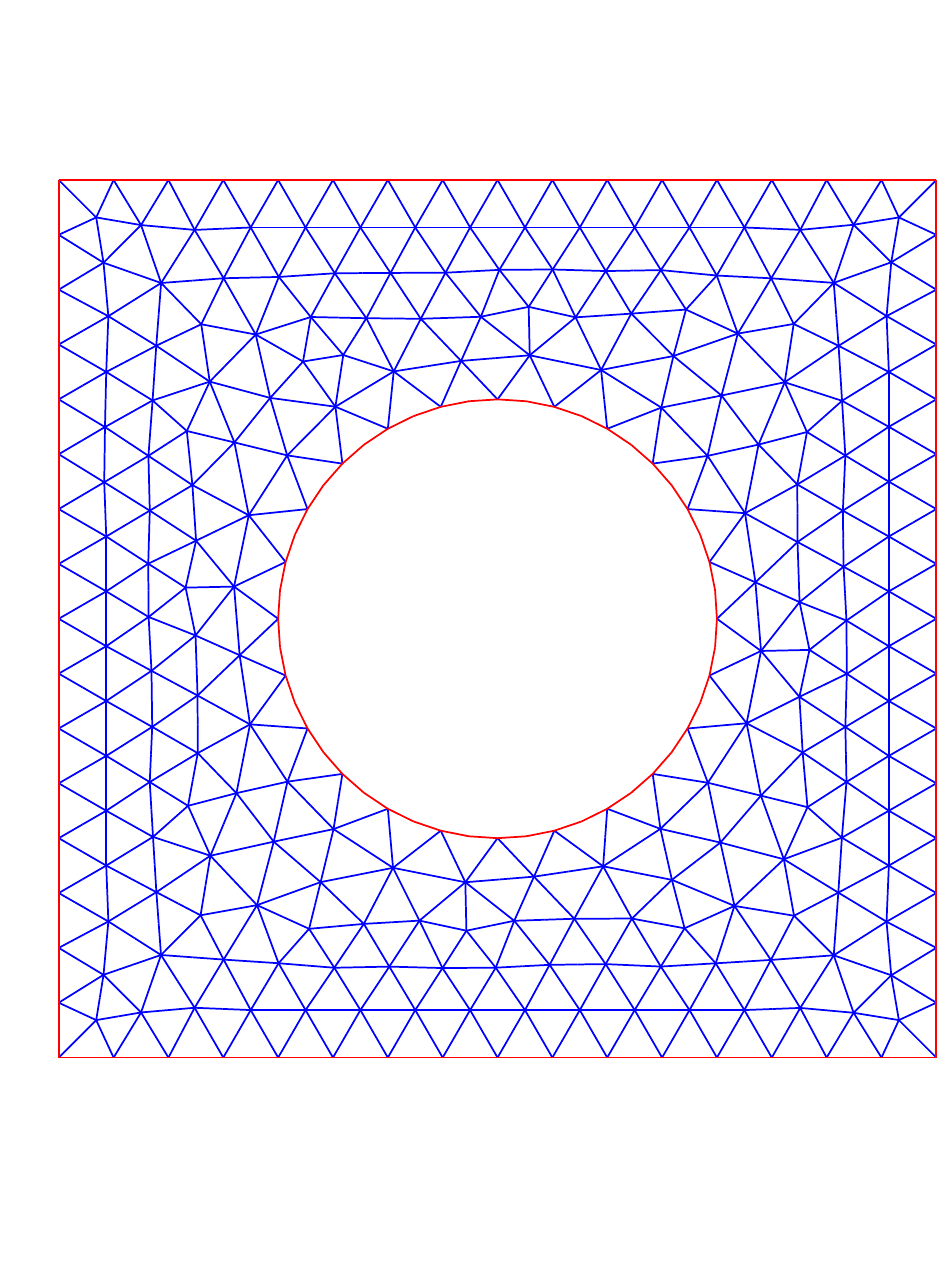}}
\subfigure[]{
\includegraphics[width=0.23\textwidth, trim= 6cm 7.5cm 6cm 7.5cm, clip]{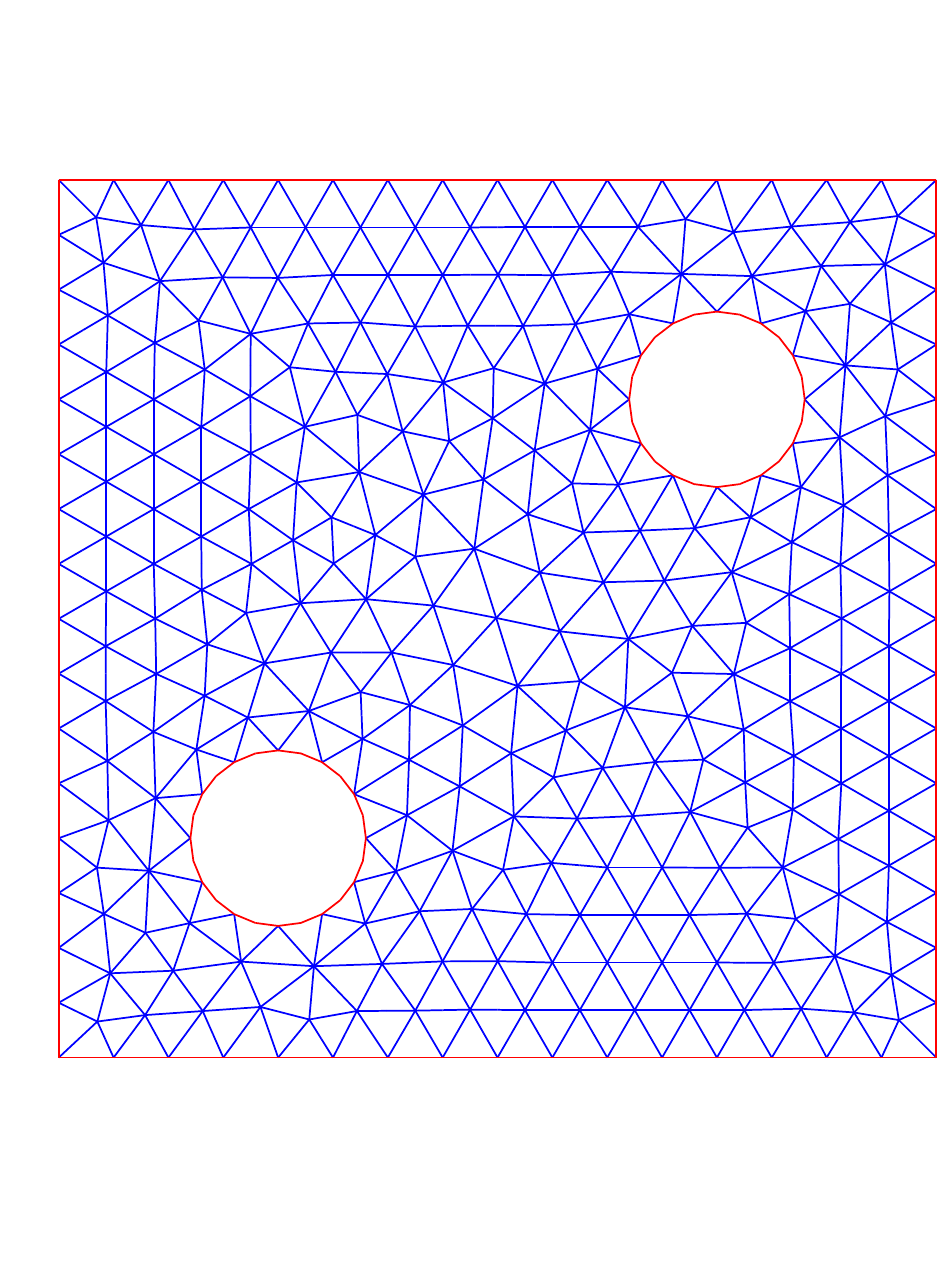}}
\subfigure[]{
\includegraphics[width=0.23\textwidth, trim= 6cm 7.5cm 6cm 7.5cm, clip]{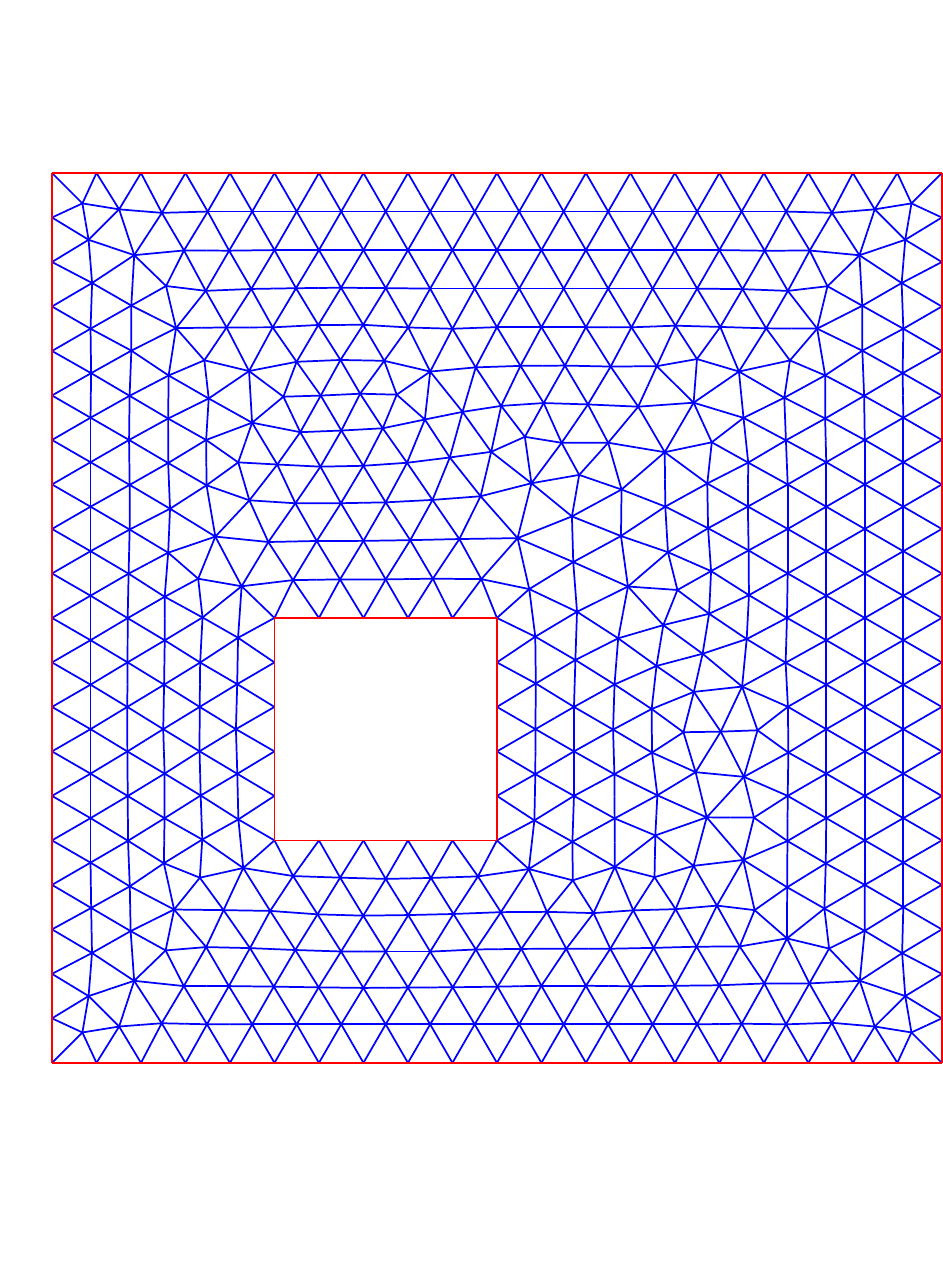}}
\subfigure[]{
\includegraphics[width=0.23\textwidth, trim= 6cm 7.5cm 6cm 7.5cm, clip]{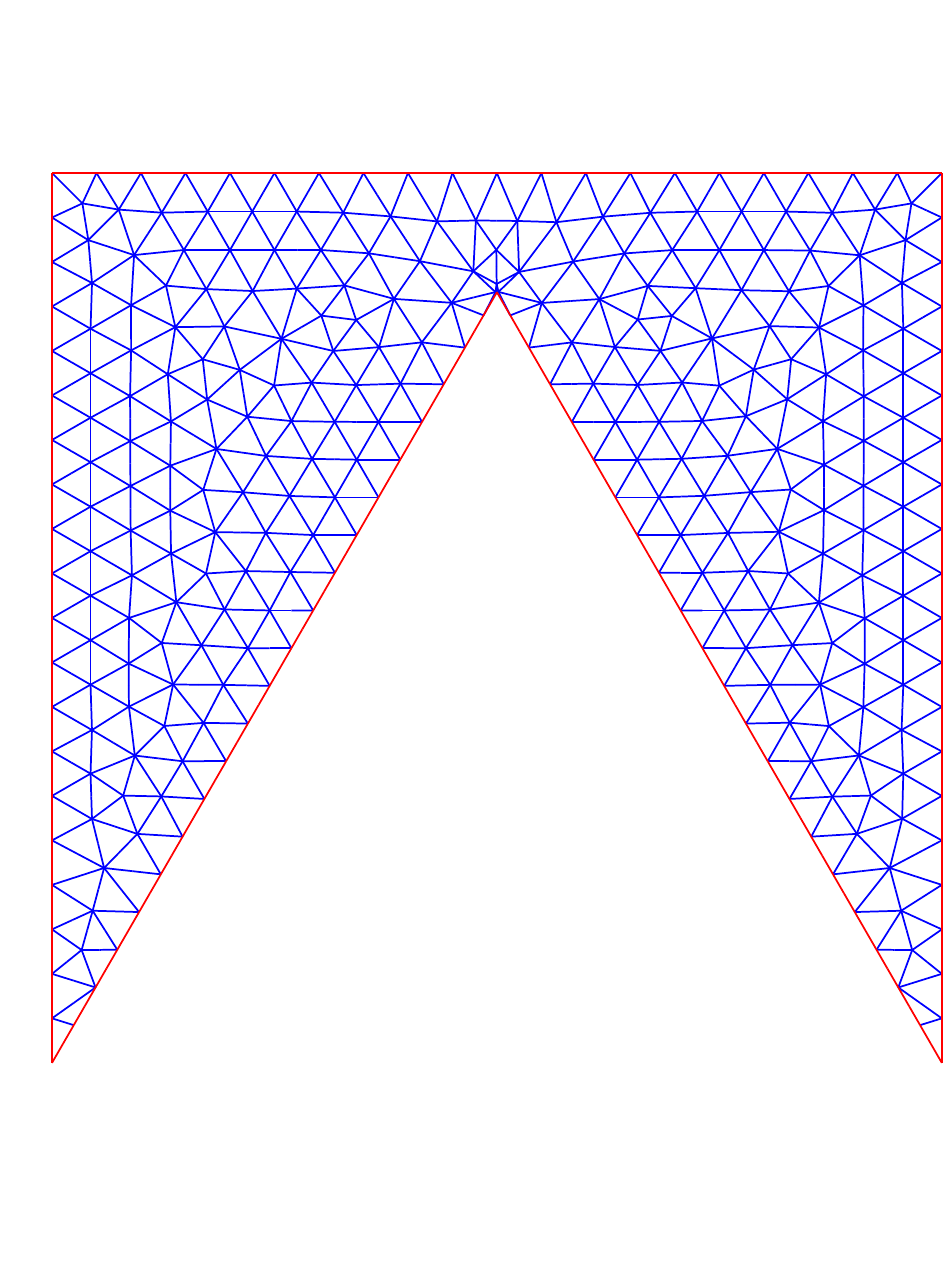}}
\caption{\review{Unstructured meshes: (a) and (e) are the source domain geometry for TL1-3 and TL4, respectively, while (b), (c), (d) and (f) are the target domains for TL1, TL2, TL3 and TL4, respectively. For the elasticity problem, (g) is the geometry of the source domain and (h) and (i) are the geometry of the target domains. Additionally, (j) is the target domain for TL10, where (i) is the source domain.}}
\label{fig:meshes}
\end{figure}

\textbf{Darcy flow and elasticity problems}
\noindent \\
All unstructured meshes for the Darcy flow and elasticity problem for the various transfer learning scenarios are shown in Figure \ref{fig:meshes}.

\begin{itemize}

\item {\bf Equilateral, right-angled triangle and triangular domain with a notch (TL1-TL3 target):} The Dirichlet boundary conditions are imposed on all boundaries i.e., $h(x, y) = 0$, for all boundary points. We employ $2295$, $1200$, $2295$ unstructured meshes in our simulations for the equilateral, the right-angled triangle and the triangular domain with a notch, respectively. For training DeepONet on the target domain, we obtain solutions for $2100$ different permeability fields, $2000$ of them are used for training the operator networks, and utilize the remaining for testing.  

\item {\bf Square domain with a vertical notch (TL4 source):} The Dirichlet boundary conditions are imposed on all boundaries i.e., $h(x, y) = 0$, for all boundary points. We employ $1538$ unstructured meshes in our simulations for the domain with a vertical notch of width, $1e-4$. For training DeepONet on the target domain, we obtain solutions for $2100$ different permeability fields, $N_t=2000$ of them are used for training the operator networks, and the remaining are used for testing. 

\item {\bf Square domain with one centered circular internal boundary (TL5-TL6 source):} We utilize the same Gaussian processes to generate the loading conditions for the right boundary here and then employ $1050$ unstructured meshes in our simulations. We generate $2000$ solutions with different boundary conditions and use $N_s=1900$ of them for training and $N^s_{\text{test}}$ for testing. The left edge is considered fixed ($u(x=0,y)=0, v(x=0,y)=0$).

\item {\bf Square domain with two circular internal boundaries in the anti-diagonal (TL5 target):} We utilize the same Gaussian processes to generate the loading conditions for the right boundary here, and then employ $1183$ unstructured meshes in our simulations. Similarly, we generate $2000$ solutions with different boundary conditions and use $N_t=1900$ of them for training and $N^t_{\text{test}}=100$ for testing.

\item {\bf Square domain with square internal boundary and irregular polygon (TL6, TL10 target and TL11):} We utilize the same Gaussian processes to generate the loading conditions for the right boundary here, and then employ $1816$ and $1166$ unstructured meshes, respectively in our simulations. Similarly, we generate $2000$ solutions with different boundary conditions and use $N_t=1900$ of them for training and $N^t_{\text{test}}=100$ for testing. For TL10, the triangular edge and the left is considered fixed. For TL6 and TL11 (source), the left edge is fixed, while for TL11 (target), the edges of the cutout are fixed.
\end{itemize}

\noindent
\textbf{Brusselator diffusion-reaction system} \\
In this problem, we aim to learn the mapping from initial concentration $h_2(x,y)$ to the evolved concentration $v(x,y,t)$, where $t>0$. The initial concentration $h_2(x,y)$ is modeled as a Gaussian random field.
\begin{equation}
\label{eq:A1:normal}
    h_2(\mathsf{x}) \sim \mathcal{GP}(h_2(\mathsf{x})\vert\mu(\mathsf{x}), \text{Cov}(\mathsf{x},\mathsf{x}')),
\end{equation}
where $\mu(\mathsf{x})$ and $\text{Cov}(\mathsf{x},\mathsf{x}')$ are the mean and covariance functions, respectively. For simplicity, we set $\mu(\mathsf{x})=0$, while the covariance matrix is given by the squared exponential kernel as

\begin{equation}
\label{eq:A2:cov}
    \text{Cov}(\mathsf{x},\mathsf{x}') = \sigma^2 \text{exp} \Bigg( - \frac{\|x - x'\|^2_2}{2\ell_x^2} - \frac{\|y - y'\|^2_2}{2\ell_y^2}\Bigg),
\end{equation}
where $\ell_x$, $\ell_y$ are the correlation length scales along the $x$ and $y$ spatial directions, respectively. To generate realizations of the input stochastic field, we employ the truncated Karhunen-Lo\'eve expansion (KLE). Datasets were generated using the \textit{py-pde} Python package that can be found in \url{https://github.com/zwicker-group/py-pde}.

\review{For the generation of data, we consider different initial concentrations of reactant $B$, as well as different KLE parameters to be used for training and testing on in- and out-of-distribution data, corresponding to the source and the target domains. These parameter sets are presented in Table \ref{table:datasets_brusselator}. }

\begin{table}[ht!]
\footnotesize
\caption{Model and input/output parameters for the source and target domains for the Brusselator dynamics.}
\centering
\begin{tabular}{c c c c c c c}
\toprule
\multirow{2}{*}{Data} & \multirow{2}{*}{$b=\{B\}$} & \multirow{2}{*}{$n_t$} & \multicolumn{3}{c}{KLE parameters} \\ \cmidrule(l){4-6} 
& & & $l_x$ & $l_y$ & $\sigma^2$ \\
\toprule 
Source &  &  & $0.30$ & $0.40$ & $0.15$ \\
OOD$_1$  & $2.2$ & $10$ & $0.11$ & $0.15$ & $0.15$ \\ 
OOD$_2$ &  & & $0.35$ & $0.45$ & $0.15$ \\ \hdashline
Target for TL7 & & &  $0.30$ & $0.40$ & $0.15$ \\ 
OOD$_1$ & $1.7$ & $10$ & $0.11$ & $0.15$ & $0.15$\\
OOD$_2$ & & & $0.35$ & $0.45$ & $0.15$  \\ \hdashline
Target for TL8 & & & $0.30$ & $0.40$ & $0.15$ \\
OOD$_1$ & $3.0$ & $10$ & $0.11$ & $0.15$ & $0.15$ \\
OOD$_2$ & & & $0.35$ & $0.45$ & $0.15$ \\ 
\bottomrule
\end{tabular}
\label{table:datasets_brusselator}
\end{table}
\section{Additional experiments}
\label{sec:add-experiments} 
\bigbreak
\noindent
\textbf{Darcy flow} 

\noindent
\review{For the Darcy flow application, we consider another problem where the goal is to map the forcing term $g(\boldsymbol x)$ in Equation \ref{eq:Darcy} which is now modeled as a Gaussian random field, to the model response $h(\boldsymbol x)$, i.e.\ $\mathcal G_{\theta}: g(\boldsymbol x) \rightarrow h(\boldsymbol x)$, while $K(\boldsymbol x)$ remains fixed. The following transfer learning task is considered:
\begin{itemize}
    \item \textbf{TL9:} Transfer learning from a square domain (source) to a triangular domain with a notch (target).
\end{itemize}
The above TL scenario is considered challenging as the external boundaries of the target domain differ significantly to those of the source. We note that this problem differs from TL4, as the neural operator aims to learn the mapping between the forcing term $g(\boldsymbol x)$ to the model response. For the source model, we generate $N_s=2{\small,}000$ train and $N^s_{\text{test}}=200$ test data. In addition, we generate $N_t=2{\small,}000$, $N^t_{\text{test}}=200$ train and test target data for training the target model. The results for TL9 are shown in Table \ref{table:TL9}. Training DeepONet on the target domain resulted to a relative $L_2$ norm error (\%) of $1.7 \pm 0.33 $ while TL-DeepONet to a $4.90 \pm 0.17$ error for the same number of training data, similar to the predictive accuracy in problem TL3 (see Table \ref{table:TL1-4}), which indicates that the choice of the input function does not result in a more challenging problem.}
\begin{table}[ht!]
\footnotesize
\caption{\review{Relative $L_2$ error and training cost in seconds ($s$) for the Darcy flow transfer learning problem (TL9).}}
\centering
\begin{tabular}{c c c c c c}
\toprule
& \multirow{2}{*}{$N_t$}  & \multicolumn{2}{c}{$L_2$ ($\%$)} & \multicolumn{2}{c}{$L_2$ ($\%$) (without $\mathcal L_{\text{CEOD}})$}\\  \cmidrule(l){3-4}  \cmidrule(l){5-6} 
& & $h(\boldsymbol x)$ &  time ($s$)  & $h(\boldsymbol x)$ &  time ($s$)\\
 \toprule
\begin{tabular}{@{}c@{}}Training DeepONet \\ (source)\end{tabular} & $2{\small,}000$  &  $2.25 \pm 0.25$  & $17{\small,}250$ &  $2.25 \pm 0.25$  & $17{\small,}250$\\ \hdashline
\begin{tabular}{@{}c@{}}Training DeepONet \\ (target)\end{tabular} & $2{\small,}000$  &  $1.7 \pm 0.33$ & $16{\small,}830$ &  $1.7 \pm 0.33$ & $16{\small,}830$\\ \hdashline
\multirow{7}{*}{\begin{tabular}{@{}c@{}}Training \\ TL-DeepONet\end{tabular}} & $5$  &  $69.38 \pm 1.60$  & $20$  &  $108.32 \pm 2.10$  & $108$\\  
& $20$ & $51.58 \pm 1.32$  & $115$ & $106.66 \pm 1.66$  & $118$\\  
& $50$  & $49.2 \pm 6.76$  & $300$ & $81.98 \pm 1.32$  & $128$\\
& $100$  & $35.77 \pm 2.06$  & $351$  & $76.58 \pm 1.32$  & $135$\\
& $150$  & $12.81 \pm 0.46$  & $380$ & $64.83 \pm 1.32$  & $142$\\
& $200$  & $11.11 \pm 1.25$  & $408$ & $61.11 \pm 1.32$  & $165$\\
& $250$  & $5.87 \pm 0.34$  & $562$ & $52.18 \pm 1.32$  & $170$\\
& $2{\small,}000$ & $4.90 \pm 0.17$ & $713$ & $16.66 \pm 1.32$  & $177$\\
\bottomrule
\end{tabular}
\label{table:TL9}
\end{table}
\bigbreak
\noindent
\textbf{Brusselator diffusion-reaction system} 

\noindent
\review{To evaluate the model performance on extrapolation we test all models (DeepONet on source and target domain and TL-DeepONet) on two out-of-distribution datasets (OOD$_1$ and OOD$_2$). The details regarding the generation of these datasets are given in Table \ref{table:datasets_brusselator}. In Table \ref{table:OOD}, all results are shown in detail. The reported values correspond to the mean relative $L_2$ norm error $\pm$ one standard deviation based on five independent runs, each time with different random seed. We note that dataset OOD$_1$ is more challenging as it corresponds to smaller length scale KLE parameters and thus it results in input fields with greater complexity. We observe that TL-DeepONet performs satisfactorily for both OOD$_1$ and OOD$_2$ even when a few data are available ($<250$).}

\begin{table}[ht!]
\footnotesize
\caption{\review{Relative $L_2$ error (\%) on two out-of-distribution datasets (OOD$_1$, OOD$_2$) for Brusselator transfer learning problems TL7 \& TL8.}}
\centering
\begin{tabular}{c c c c c c}
\toprule
& \multirow{2}{*}{\begin{tabular}{@{}c@{}}\# of training \\ data ($N_t$)\end{tabular}}  & \multicolumn{2}{c}{TL7} & \multicolumn{2}{c}{TL8} \\  \cmidrule(l){3-4}  \cmidrule(l){5-6}  & & OOD$_1$  & OOD$_2$ & OOD$_1$  & OOD$_2$  \\
 \toprule
\begin{tabular}{@{}c@{}}Training \\ DeepONet \\ (source) \end{tabular} & $800$  &  $2.27 \pm 0.12$ & $1.66 \pm 0.13$ & $2.27 \pm 0.12$ & $1.66 \pm 0.13$   \\ \hdashline
\begin{tabular}{@{}c@{}}Training \\ DeepONet \\ (target)\end{tabular} & $800$  &  $2.10 \pm 0.29$ & $1.57 \pm 0.05$ & $4.35 \pm 0.12$  & $3.30 \pm 0.15$  \\ \hdashline
\multirow{7}{*}{\begin{tabular}{@{}c@{}}Training \\ TL-DeepONet\end{tabular}} & $5$  &  $27.2 \pm 1.70$ & $23.86 \pm 1.54$ & $59.39 \pm 5.9$ & $51.67 \pm 5.66$  \\  
& $20$ & $19.10 \pm 1.36$ & $14.87 \pm 1.40$ & $19.11 \pm 1.37$ & $17.56 \pm 1.42$  \\  
& $50$  & $4.56 \pm 0.06$ & $2.72 \pm 0.02$ & $17.39 \pm 1.17$ & $15.93 \pm 1.18$  \\
& $100$  & $3.75 \pm 0.22$ & $2.50 \pm 0.04$ & $14.85 \pm 0.53$ & $12.05 \pm 0.45$  \\
& $150$  & $3.30 \pm 0.08$ & $2.41 \pm 0.09$ & $13.24 \pm 0.96$ & $11.17 \pm 0.88$  \\
& $200$  & $2.96 \pm 0.05$ & $2.35 \pm 0.05$ & $8.04 \pm 0.14$ & $7.54 \pm 0.76$  \\
& $250$  & $2.93 \pm 0.05$ & $2.21 \pm 0.07$ & $5.69 \pm 0.36$ & $5.94 \pm 0.33$ \\
& $800$ & $2.33 \pm 0.01$ & $2.12 \pm 0.02$  & $5.47 \pm 0.28$ & $5.78 \pm 0.42$  \\
\bottomrule
\end{tabular}
\label{table:OOD}
\end{table}
\bigbreak
\noindent
\textbf{Elasticity model} 

\begin{figure}[ht!]
\begin{center}
\includegraphics[width=0.85\textwidth]{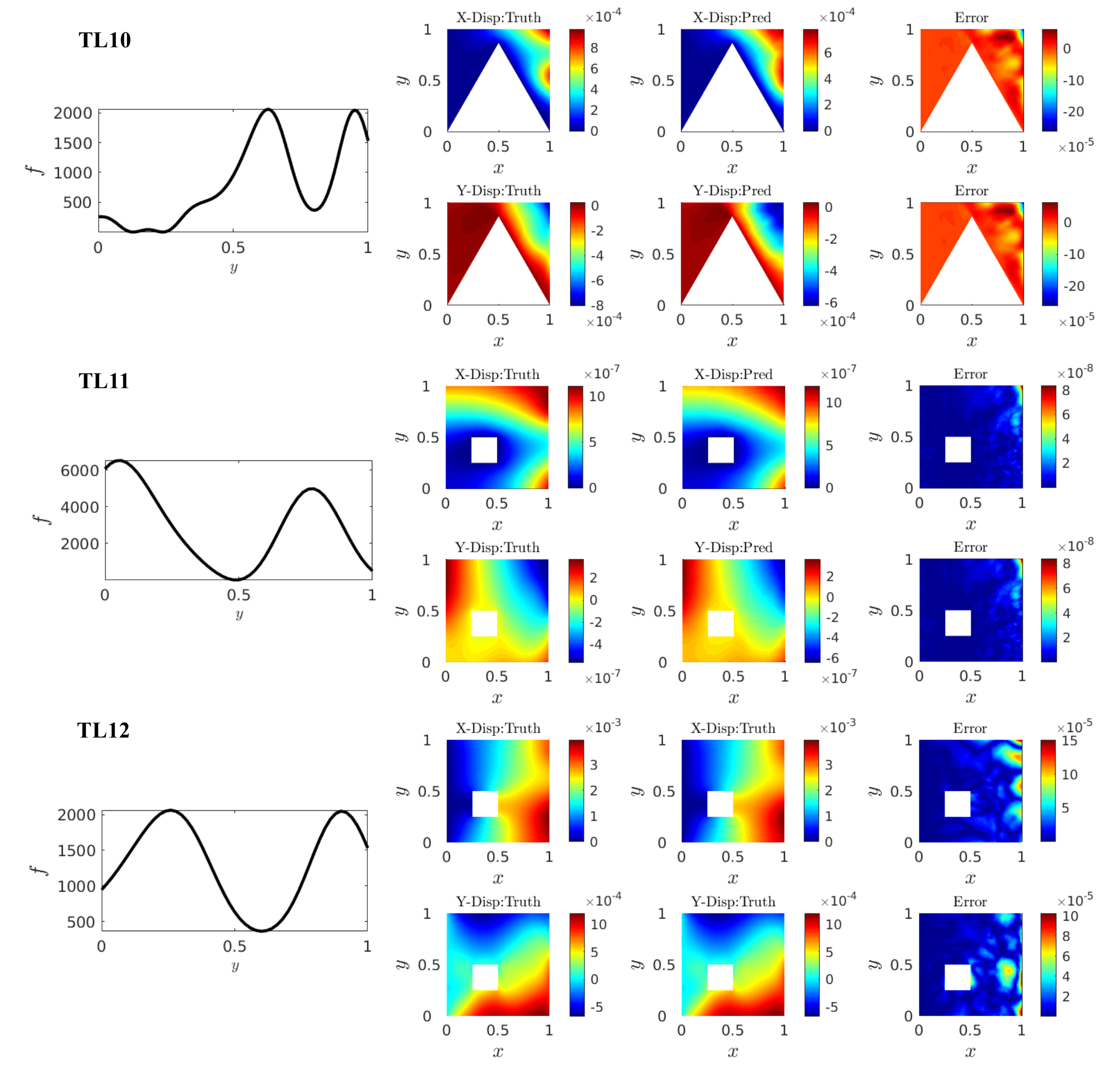}
\caption{\review{Representative results for the transfer learning problems TL10 and TL11 for which DeepONet takes as input the loading condition applied on the right edge of the plate and outputs the displacement field.}}
\label{fig:results-TL10-11-12}
\end{center}
\end{figure}

\noindent
\review{To test the limitations of the proposed approach we consider two additional transfer learning scenarios:
\begin{itemize}
    \item \textbf{TL10:} Transfer learning from domain with a square internal boundary and material properties $(E_S=300\times10^5,\nu_S=0.3)$ to an irregular polygon domain (different external boundaries) and different material properties $(E_T=410\times10^3,\nu_T=0.35)$.
    \item \textbf{TL11:} Transfer learning from domain with material properties $(E_S=300\times10^5,\nu_S=0.3)$ and left boundary fixed to domain with different material properties $(E_T=410\times10^3,\nu_T=0.35)$ and boundaries of the square cutout fixed (for both source and target, the geometry corresponds to square with square internal boundary).
    \item \textbf{TL12:} Transfer learning from a domain with a centered circular internal boundary and material properties $(E_S=300\times10^5,\nu_S=0.3)$ to a domain with a square internal boundary and different material properties $(E_T=410\times10^3,\nu_T=0.35)$.
\end{itemize}
The results for the above TL problems are shown in Tables \ref{table:TL10}, \ref{table:TL11}, and \ref{table:TL12}. For TL10, the main challenge stems from the different internal and external boundaries and the material properties between the two domains. From the relatively higher error of TL-DeepONet (Table \ref{table:TL10}) we observe that the significant differences between the source and target domains cause the model accuracy to deteriorate as the transfer of features extracted from the source model is insufficient during the training of TL-DeepONet. These results demonstrate the limitations of the proposed transfer learning approach. From Table \ref{table:TL11}, we found that when both the material properties and boundary conditions between source and target are modified, TL-DeepONet performs very well even for a small number of training samples. From Table \ref{table:TL12}, we found that TL-DeepONet can efficiently handle the transition from a smooth circular cutout to a square cutout with different material properties. In TL12, we present a case where there is change in the geometric domain and a small change in the material properties (e.g. $\nu_S = 0.3$ and $\nu_T = 0.35$). We found that TL-DeepONet can be implemented to obtain accurate predictions on the target domain. TL12 can be considered to be a modified case of TL6, with a smaller increase in the Poisson ratio. The magnitude of the errors are roughly doubled in TL6, where we have $\nu = 0.45$. Therefore, the major errors in transfer learning for TL6/TL12 come not from domain adaptation (despite the non-smoothness) but rather from large change in Poisson's ratio value. We also show representative results for each task in Figure \ref{fig:results-TL10-11-12}. 
}
\begin{table}[ht!]
\small
\caption{\review{Relative $L_2$ error and training cost in seconds ($s$) for the elasticity transfer learning problem (TL10).}}
\centering
\footnotesize
\begin{tabular}{c c c c c}
\toprule
& \multirow{2}{*}{$N_t$}  & \multicolumn{3}{c}{$L_2$ ($\%$)} \\  \cmidrule(l){3-5} & & $u(\boldsymbol x)$ & $v(\boldsymbol x)$ & time ($s$)  \\
 \toprule
\begin{tabular}{@{}c@{}}Training DeepONet \\ (source)\end{tabular} & $1{\small,}900$  &  $2.30 \pm 0.49$ & $3.22 \pm 0.48$ & $10{\small,}060$ \\ \hdashline
\begin{tabular}{@{}c@{}}Training DeepONet \\ (target)\end{tabular} & $1{\small,}900$  &  $2.72 \pm 0.26$ & $1.92 \pm 0.41$ & $11{\small,}750$ \\ \hdashline
\multirow{7}{*}{\begin{tabular}{@{}c@{}}Training \\ TL-DeepONet \end{tabular}} & $5$  &  $60.28 \pm 1.95$ & $57.48 \pm 2.54$ & $18$  \\  
& $20$ & $29.14 \pm 0.31$ & $21.42 \pm 3.20$ & $148$ \\  
& $50$  & $16.4 \pm 2.01$ & $18.72 \pm 1.18$ & $25$ \\
& $100$  & $11.37 \pm 0.34$ & $14.15 \pm 0.96$ & $44$  \\
& $150$  & $9.66 \pm 0.26$ & $11.95 \pm 0.61$ & $116$ \\
& $200$  & $5.52 \pm 0.20$ & $10.94 \pm 0.20$ & $384$ \\
& $250$  & $4.08 \pm 0.06$ & $9.18 \pm 0.14$ & $480$ \\
& $1{\small,}900$ & $3.82 \pm 0.20$ & $7.89 \pm 0.21$ &  $512$ \\
\bottomrule
\end{tabular}
\label{table:TL10}
\end{table}
\begin{table}[ht!]
\small
\caption{\review{Relative $L_2$ error and training cost in seconds ($s$) for the elasticity transfer learning problem (TL11).}}
\centering
\footnotesize
\begin{tabular}{c c c c c}
\toprule
& \multirow{2}{*}{$N_t$}  & \multicolumn{3}{c}{$L_2$ ($\%$)} \\  \cmidrule(l){3-5} & & $u(\boldsymbol x)$ & $v(\boldsymbol x)$ & time ($s$)  \\
 \toprule
\begin{tabular}{@{}c@{}}Training DeepONet \\ (source)\end{tabular} & $1{\small,}900$  &  $2.30 \pm 0.49$ & $3.22 \pm 0.48$ & $10{\small,}060$ \\ \hdashline
\begin{tabular}{@{}c@{}}Training DeepONet \\ (target)\end{tabular} & $1{\small,}900$  &  $2.72 \pm 0.26$ & $1.92 \pm 0.41$ & $11{\small,}750$ \\ \hdashline
\multirow{2}{*}{\begin{tabular}{@{}c@{}}Training TL-DeepONet \end{tabular}} 
& $250$  & $4.08 \pm 0.06$ & $4.18 \pm 0.14$ & $480$ \\
& $1{\small,}900$ & $3.82 \pm 0.20$ & $3.89 \pm 0.21$ &  $512$ \\
\bottomrule
\end{tabular}
\label{table:TL11}
\end{table}
\begin{table}[ht!]
\small
\caption{\review{Relative $L_2$ error and training cost in seconds ($s$) for the elasticity transfer learning problem (TL12).}}
\centering
\footnotesize
\begin{tabular}{c c c c c}
\toprule
& \multirow{2}{*}{$N_t$}  & \multicolumn{3}{c}{$L_2$ ($\%$)} \\  \cmidrule(l){3-5} & & $u(\boldsymbol x)$ & $v(\boldsymbol x)$ & time ($s$)  \\
 \toprule
\begin{tabular}{@{}c@{}}Training DeepONet \\ (source)\end{tabular} & $1{\small,}900$  &  $2.30 \pm 0.49$ & $3.22 \pm 0.48$ & $10{\small,}060$ \\ \hdashline
\begin{tabular}{@{}c@{}}Training DeepONet \\ (target)\end{tabular} & $1{\small,}900$  &  $2.98 \pm 0.14$ & $3.9 \pm 0.50$ & $11{\small,}105$ \\ \hdashline
\multirow{2}{*}{\begin{tabular}{@{}c@{}}Training TL-DeepONet \end{tabular}} 
& $250$  & $5.36 \pm 0.06$ & $5.62 \pm 0.20$ & $382$ \\
& $1{\small,}900$ & $3.92 \pm 0.20$ & $4.26 \pm 0.31$ &  $510$ \\
\bottomrule
\end{tabular}
\label{table:TL12}
\end{table}

\bigbreak
\noindent
\textbf{Burgers' equation} 

\begin{figure}[ht!]
\begin{center}
\includegraphics[width=1\textwidth]{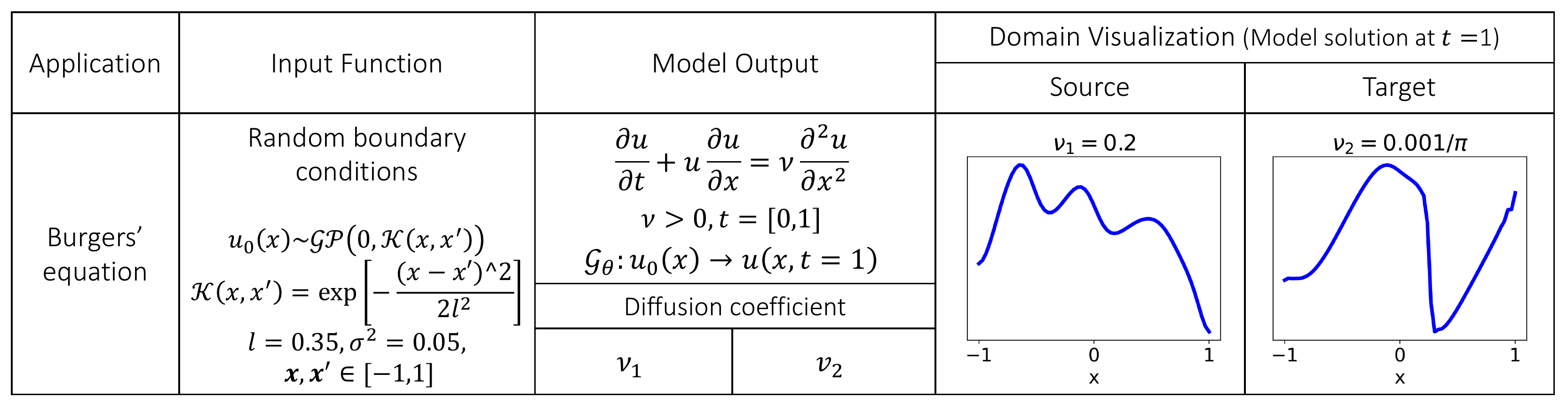}
\caption{\review{A schematic representation of the operator learning problem considered for the Burgers' equation. The input/output functions and representative plots of the source and the target domains are shown.}}
\label{fig:application-burgers}
\end{center}
\end{figure}

\begin{figure}[ht!]
\begin{center}
\includegraphics[width=0.85\textwidth]{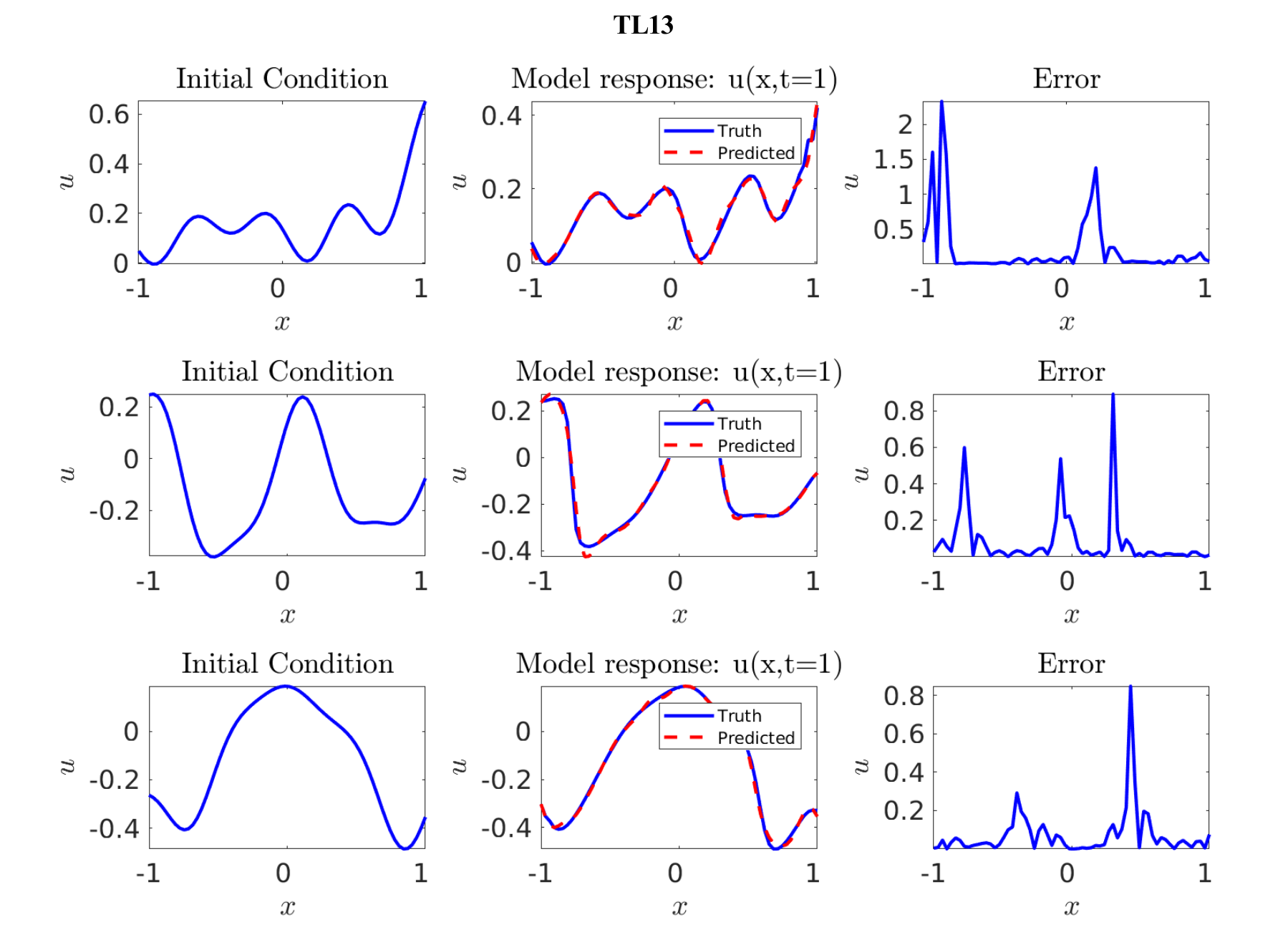}
\caption{Three representative results for Burgers' equation transfer learning problem TL13 for which DeepONet takes as input the random boundary conditions and outputs the model response at the last simulation time step.Error plots represent the point-wise error computed as $\lvert \frac{f(\mathbf{x}^{T})- {\mathbf{y}^T}}{{\mathbf{y}^T}}\rvert $, where $\mathbf{y}^T$, $f(\mathbf{x}^{T})$ is the reference response and the model prediction, respectively.}
\label{fig:results-TL13}
\end{center}
\end{figure}

\noindent
\review{To further evaluate the performance of TL-DeepONet in non-linear problems, we consider the 1D viscous Burgers' equation, a nonlinear parabolic PDE of the following form
\begin{equation}
    \frac{\partial u}{\partial t} + u\frac{\partial u}{\partial x}=  \nu \frac{\partial^2 u}{\partial x^2}, \ \ \ x \in [-1,1], \ t \in [0,1]
\label{eq:burgers}
\end{equation}
where $u(x,t)$ represents the velocity field of the fluid and $\nu >0$ is the diffusion coefficient. We train DeepONet to learn the mapping between the initial condition $u(x,t=0)=u_0(x)$ and the Burgers' response at the final simulation step, i.e.\ $\mathcal G_{\theta}: u_0(x) \rightarrow u(x, t=1)$. The initial field $u_0(x)$ is modeled as a Gaussian random field. We generate random realizations based on KLE with parameters $l_x=0.35$ and $\sigma^2=0.05$. To solve the above system we use a finite difference scheme with spatial discretization of $\Delta x = 0.03125$ and a temporal discretization of $\Delta t = 0.001$, and we keep the model solution at the last simulation step. The following transfer learning problem is considered (see also Figure \ref{fig:application-burgers}):
\begin{itemize}
    \item \textbf{TL13:} Transfer learning from diffusion coefficients $\nu_1=0.2$ to $\nu_2=0.001/\pi$. 
\end{itemize}
In the above scenario, data for the source domain have been generated with $\nu_1=0.2$, which results in a relatively smooth response, while for the target domain with $\nu_2=0.001/\pi$ sharp non-linearities appear in the model response. For the source model, we generate $N_s=2{\small,}000$ train and $N^s_{\text{test}}=200$ test data. In addition, we generate $N_t=2{\small,}000$, $N^t_{\text{test}}=200$ train and test target data for training the target model. The results of problem TL13 are presented in Table \ref{table:TL13}. The reported results represent the mean $\pm$ one standard deviation based on five independent runs with different random seed. In Table \ref{table:TL13}, we observe that TL-DeepONet results in satisfactory performance, however, only when sufficient labeled data are considered. This is an expected result, as TL-DeepONet is expected to learn the non-linearities in the model response only through the fine-tuning of the fully-connected layers, which is a challenging task. Finally, plots of three representative realizations of the initial condition with reference response and TL-DeepONet error are shown in Figure \ref{fig:results-TL13}. }
\begin{table}[ht!]
\small
\caption{\Review{Relative $L_2$ error and training cost in seconds ($s$) for the Burgers' equation transfer learning problem (TL13).}}
\centering
\footnotesize
\begin{tabular}{c c c c}
\toprule
& \multirow{2}{*}{$N_t$}  & \multicolumn{2}{c}{$L_2$ ($\%$)} \\  \cmidrule(l){3-4} & & $u(x,t=1)$ &  time ($s$)  \\
 \toprule
\begin{tabular}{@{}c@{}}Training DeepONet \\ (source)\end{tabular} & $2{\small,}000$  &  $2.40 \pm 0.11$  & $1{\small,}470$ \\ \hdashline
\multirow{7}{*}{\begin{tabular}{@{}c@{}}Training DeepONet \\ (target)\end{tabular}} 
& $5$  &  $85.26 \pm 1.73$  & $1220$  \\  
& $20$ & $35.66 \pm 0.67$  & $1220$ \\  
& $50$  & $23.00 \pm 0.41$  & $1220$ \\
& $100$  & $17.92 \pm 0.33$  & $1230$  \\
& $150$  & $14.82 \pm 0.33$  & $1230$ \\
& $200$  & $12.18 \pm 0.27$  & $1230$ \\
& $250$  & $10.85 \pm 0.23$  & $1240$ \\
& $2{\small,}000$  &  $2.42 \pm 0.13$ & $1{\small,}310$ \\ \hdashline
\multirow{7}{*}{\begin{tabular}{@{}c@{}}Training \\ TL-DeepONet \end{tabular}} & $5$  &  $150.64 \pm 17.78$  & $8$  \\  
& $20$ & $29.63 \pm 1.46$  & $21$ \\  
& $50$  & $15.26 \pm 0.47$  & $46$ \\
& $100$  & $10.56 \pm 0.77$  & $55$  \\
& $150$  & $5.11 \pm 0.51$  & $61$ \\
& $200$  & $4.83 \pm 0.44$  & $75$ \\
& $250$  & $3.57 \pm 0.22$  & $79$ \\
& $2{\small,}000$ & $3.42 \pm 0.22$ & $98$ \\
\bottomrule
\end{tabular}
\label{table:TL13}
\end{table}

\begin{thebibliography}{50}

\bibitem{chen2018neural} Chen, R. T. and Rubanova, Y. and Bettencourt, J. and Duvenaud, D.K., Neural Ordinary Differential Equations. Advances in Neural Information Processing Systems 31 (2018).

\bibitem{raissi2019physics} Raissi, M., Perdikaris, P., Karniadakis, G.E.: Physics-informed neural
networks: A deep learning framework for solving forward and inverse problems involving nonlinear partial differential equations. Journal of Computational Physics 378, 686–707 (2019).

\bibitem{li2021fourier} Li, Z., Kovachki, N.B., Azizzadenesheli, K., liu, B., Bhattacharya, K., Stuart, A., Anandkumar, A.: Fourier Neural Operator for Parametric Partial Differential Equations. In: In Proceedings of the International Conference on Learning Representations (2021).

\bibitem{lu2021learning} Lu, L., Jin, P., Pang, G., Zhang, Z., Karniadakis, G.E.: Learning nonlinear operators via DeepONet based on the universal approximation theorem of operators. Nature Machine Intelligence 3(3), 218–229 (2021).

\bibitem{chatterjee2021robust} Chatterjee, T., Chakraborty, S., Goswami, S., Adhikari, S., Friswell, M.I.: Robust topological designs for extreme metamaterial micro-structures. Scientific Reports 11(1), 1–14 (2021).

\bibitem{olivier2021bayesian} Olivier, A., Shields, M.D., Graham-Brady, L.: Bayesian neural networks for uncertainty quantification in data-driven materials modeling. Computer Methods in Applied Mechanics and Engineering 386, 114079 (2021).

\bibitem{niu2020decade} Niu, S., Liu, Y., Wang, J., Song, H.: A Decade Survey of Transfer Learning
(2010–2020). IEEE Transactions on Artificial Intelligence 1(2), 151–166 (2020).

\bibitem{gao2018deep} Gao, Y., Mosalam, K.M.: Deep Transfer Learning for Image-Based Structural Damage Recognition. Computer-Aided Civil and Infrastructure Engineering 33(9), 748–768 (2018).

\bibitem{yang2021image} Yang, X., Zhang, Y., Lv, W., Wang, D.: Image recognition of wind turbine
blade damage based on a deep learning model with transfer learning and an ensemble learning classifier. Renewable Energy 163, 386–397 (2021)

\bibitem{ruder2019transfer} Ruder, S., Peters, M.E., Swayamdipta, S., Wolf, T.: Transfer Learning
in Natural Language Processing. In: Proceedings of the 2019 Conference of the North American Chapter of the Association for Computational Linguistics: Tutorials, 15–18 (2019).

\bibitem{zhang2021combining} Zhang, S., Chen, M., Chen, J., Li, Y.-F., Wu, Y., Li, M., Zhu, C.: Combining cross-modal knowledge transfer and semi-supervised learning for speech emotion recognition. Knowledge-Based Systems 229, 107340 (2021).

\bibitem{zhuang2020comprehensive} Zhuang, F., Qi, Z., Duan, K., Xi, D., Zhu, Y., Zhu, H., Xiong, H., He, Q.: A Comprehensive Survey on Transfer Learning. Proceedings of the IEEE 109(1), 43–76 (2020).

\bibitem{certo2016sample} Certo, S.T., Busenbark, J.R., Woo, H.-s., Semadeni, M.: Sample selection
bias and Heckman models in strategic management research. Strategic Management Journal 37(13), 2639–2657 (2016).

\bibitem{chen2021representation} Chen, X., Wang, S., Wang, J., Long, M.: Representation Subspace Distance for Domain Adaptation Regression. In: Proceedings of the 38th International Conference on Machine Learning, 1749–1759 (2021).

\bibitem{pardoe2010boosting} Pardoe, D., Stone, P.: Boosting for regression transfer. In: Proceedings of the 27th International Conference on Machine Learning, 863–870 (2010).

\bibitem{wang2014active} Wang, X., Huang, T.-K., Schneider, J.: Active Transfer Learning under
Model Shift. In: Proceedings of the 31st International Conference on Machine Learning, 1305–1313 (2014).

\bibitem{du2017hypothesis} Du, S.S., Koushik, J., Singh, A., P{\'o}czos, B.: Hypothesis Transfer Learning via Transformation Functions. Advances in Neural Information Processing Systems 30 (2017).

\bibitem{zhang2013domain} Zhang, K., Sch{\"o}olkopf, B., Muandet, K., Wang, Z.: Domain Adaptation
under Target and Conditional Shift. In: Proceedings of the International Conference on Machine Learning, 819–827 (2013).

\bibitem{chen2020transfer} Chen, G., Li, Y., Liu, X.: Transfer Learning Under Conditional Shift
Based on Fuzzy Residual. IEEE Transactions on Cybernetics (2020).

\bibitem{liu2021deep} Liu, X., Li, Y., Meng, Q., Chen, G.: Deep transfer learning for conditional shift in regression. Knowledge-Based Systems 227, 107216 (2021).

\bibitem{zhang2020machine} Zhang, X., Garikipati, K.: Machine learning materials physics: Multi-
resolution neural networks learn the free energy and nonlinear elastic response of evolving microstructures. Computer Methods in Applied Mechanics and Engineering 372, 113362 (2020).

\bibitem{goswami2020transfer} Goswami, S., Anitescu, C., Chakraborty, S., Rabczuk, T.: Transfer learning enhanced physics informed neural network for phase-field modeling of fracture. Theoretical and Applied Fracture Mechanics 106, 102447 (2020).

\bibitem{desai2021one} Desai, S., Mattheakis, M., Joy, H., Protopapas, P., Roberts, S.: One-Shot Transfer Learning of Physics-Informed Neural Networks. arXiv preprint arXiv:2110.11286 (2021).

\bibitem{chen2021transfer} Chen, X., Gong, C., Wan, Q., Deng, L., Wan, Y., Liu, Y., Chen, B., Liu, J.: Transfer learning for deep neural network-based partial differential equations solving. Advances in Aerodynamics 3(1), 1–14 (2021).

\bibitem{penwarden2021physics} Penwarden, M., Zhe, S., Narayan, A., Kirby, R.M.: Physics-Informed Neural Networks (PINNs) for Parameterized PDEs: A Metalearning Approach. arXiv preprint arXiv:2110.13361 (2021).

\bibitem{wang2022mosaic} Wang, H., Planas, R., Chandramowlishwaran, A., Bostanabad, R.: Mosaic
flows: A transferable deep learning framework for solving PDEs on unseen domains. Computer Methods in Applied Mechanics and Engineering 389, 114424 (2022).

\bibitem{neyshabur2020being} Neyshabur, B., Sedghi, H., Zhang, C.: What is being transferred in transfer learning? Advances in Neural Information Processing Systems 33, 512–523 (2020).

\bibitem{tripura2022wavelet} Tripura, T., Chakraborty, S.: Wavelet neural operator: a neural operator for parametric partial differential equations. arXiv preprint arXiv:2205.02191 (2022).

\bibitem{li2020neural} Li, Z., Kovachki, N., Azizzadenesheli, K., Liu, B., Bhattacharya, K., Stuart, A., Anandkumar, A.: Neural operator: Graph kernel network for partial differential equations. arXiv preprint arXiv:2003.03485 (2020).

\bibitem{lu2021comprehensive} Lu, L., Meng, X., Cai, S., Mao, Z., Goswami, S., Zhang, Z., Karniadakis, G.E.: A comprehensive and fair comparison of two neural operators (with practical extensions) based on FAIR data. Computer Methods in Applied Mechanics and Engineering 393, 114778 (2022).

\bibitem{ahmed2019numerical} Ahmed, N., Rafiq, M., Rehman, M., Iqbal, M., Ali, M.: Numerical modeling of three dimensional Brusselator reaction diffusion system. AIP Advances 9(1), 015205 (2019).

\bibitem{lee2006estimation} Lee, Y.K., Park, B.U.: Estimation of Kullback–Leibler divergence by local likelihood. Annals of the Institute of Statistical Mathematics 58(2), 327–340 (2006).

\bibitem{yu2020measuring} Yu, S., Shaker, A., Alesiani, F., Principe, J.C.: Measuring the Discrepancy between Conditional Distributions: Methods, Properties and Applications. In: Proceedings of the 29th International Joint Conference on Artificial Intelligence, 2777–2784 (2020).

\bibitem{muandet2017kernel} Muandet, K., Fukumizu, K., Sriperumbudur, B., Sch{\"o}olkopf, B., et al.: Kernel Mean Embedding of Distributions: A Review and Beyond. Foundations and Trends in Machine Learning 10(1-2), 1–141 (2017).

\bibitem{gretton2012kernel} Gretton, A., Borgwardt, K.M., Rasch, M.J., Sch{\"o}olkopf, B., Smola, A.: A Kernel Two-Sample Test. The Journal of Machine Learning Research 13(1), 723–773 (2012).

\bibitem{song2013kernel} Song, L., Fukumizu, K., Gretton, A.: Kernel Embeddings of Conditional
Distributions: A Unified Kernel Framework for Nonparametric Inference in Graphical Models. IEEE Signal Processing Magazine 30(4), 98–111 (2013).

\bibitem{song2009hilbert} Song, L., Huang, J., Smola, A., Fukumizu, K.: Hilbert Space Embeddings
of Conditional Distributions with Applications to Dynamical Systems. In: Proceedings of the 26th Annual International Conference on Machine Learning, 961–968 (2009).

\bibitem{saxe2019information} Saxe, A.M., Bansal, Y., Dapello, J., Advani, M., Kolchinsky, A., Tracey, B.D., Cox, D.D.: On the information bottleneck theory of deep learning. Journal of Statistical Mechanics: Theory and Experiment 2019(12), 124020 (2019).

\bibitem{yosinski2014transferable} Yosinski, J., Clune, J., Bengio, Y., Lipson, H.: How transferable are features in deep neural networks? Advances in Neural Information Processing Systems 27 (2014).

\bibitem{kontolati2022influence} Kontolati, K., Goswami, S., Shields, M.D., Karniadakis, G.E.: On the influence of over-parameterization in manifold based surrogates and deep neural operators. arXiv preprint arXiv:2203.05071 (2022).

\bibitem{github-code} Kontolati, K., Goswami, S., Shields, M.D., Karniadakis, G.E., TL-DeepONet: Codes for deep transfer operator learning for partial
differential equations under conditional shift. DOI: https://doi.org/10.5281/zenodo.7195684 (2022).


\end{thebibliography}
\end{document}